\documentclass{article}

\usepackage[preprint]{neurips_2026}

\usepackage[utf8]{inputenc} 
\usepackage[T1]{fontenc}    
\usepackage{url}            
\usepackage{booktabs}       
\usepackage{amsfonts}       
\usepackage{nicefrac}       
\usepackage{microtype}      
\usepackage{xcolor}         
\usepackage{graphicx}
\usepackage{titletoc}

\definecolor{nipsblue}{rgb}{0.21,0.49,0.74}
\usepackage[pagebackref,breaklinks,colorlinks,allcolors=nipsblue]{hyperref}

\usepackage{enumitem}
\usepackage{multirow}
\usepackage{wrapfig}
\usepackage{colortbl}
\usepackage[normalem]{ulem}
\usepackage{microtype}
\usepackage{comment}
\usepackage[hang,flushmargin]{footmisc}
\usepackage{graphicx}
\usepackage{multicol}
\usepackage{float}
\usepackage{pifont}
\usepackage{nicematrix}
\usepackage{ulem}
\usepackage{diagbox} 
\usepackage{tipa}
\usepackage{diagbox} 
\usepackage{nicematrix}
\usepackage{makecell}
\usepackage{microtype}

\usepackage{nicematrix}   
\usepackage{booktabs}
\usepackage{xcolor}
\usepackage{ulem}         
\usepackage{array}        
\usepackage{relsize}

\usepackage{amsmath,amssymb}
\usepackage{amsthm}
\usepackage[most]{tcolorbox}

\setlist[itemize]{align=parleft,left=0pt,topsep=1mm,itemsep=0mm,parsep=1mm}
\definecolor{azure}{rgb}{0.0, 0.5, 1.0}
\definecolor{azure(colorwheel)}{rgb}{0.0, 0.5, 1.0}
\definecolor{nicegreen}{rgb}{0.0, 0.7, 0.1}
\definecolor{yw}{rgb}{0.01176, 0.5490, 0.5490}
\definecolor{ashblue}{rgb}{0.36, 0.54, 0.66}
\definecolor{ashgrey}{rgb}{0.7, 0.75, 0.71}
\definecolor{applegreen}{rgb}{0.55, 0.71, 0.0}
\definecolor{blue}{rgb}{0.0, 0.0, 1.0}
\definecolor{red}{rgb}{0.6, 0.0, 0.12}
\definecolor{postechred}{rgb}{0.784, 0.003, 0.313}
\definecolor{ywg}{rgb}{0.9960, 0.8984, 0.5859}
\definecolor{ballblue}{rgb}{0.13, 0.67, 0.8}
\definecolor{cornellred}{rgb}{0.7, 0.11, 0.11}
\definecolor{darkcyan}{rgb}{0.0, 0.55, 0.55}
\definecolor{CuGray}{gray}{0.9}
\definecolor{airforceblue}{rgb}{0.36, 0.54, 0.66}
\definecolor{rev}{rgb}{0.784, 0.003, 0.313}
\definecolor{pink}{cmyk}{0, 0.7808, 0.4429, 0.1412}
\definecolor{amethyst}{rgb}{0.6, 0.4, 0.8}
\definecolor{black}{rgb}{0.0, 0.0, 0.0}
\definecolor{tb3_yellow}{rgb}{0.996, 1.0, 0.6}
\definecolor{tb3_orange}{rgb}{0.980, 0.8, 0.604}
\definecolor{tb3_red}{rgb}{0.972, 0.6, 0.6}
\definecolor{dimgray}{rgb}{0.41, 0.41, 0.41}
\definecolor{brickred}{rgb}{0.8, 0.25, 0.33}
\definecolor{bleudefrance}{rgb}{0.19, 0.55, 0.91}
\definecolor{blue(ncs)}{rgb}{0.265, 0.445, 0.765}
\definecolor{blue(ryb)}{rgb}{0.01, 0.28, 1.0}
\definecolor{orange}{rgb}{1.0, 0.49, 0.0}
\definecolor{Gray}{gray}{0.88}
\definecolor{green(ncs)}{rgb}{0.0, 0.62, 0.42}
\definecolor{clova}{rgb}{0.24, 0.63, 0.33}
\definecolor{correctgreen}{rgb}{0.01, 0.702, 0.321}
\definecolor{wrongred}{rgb}{0.757, 0, 0}

\definecolor{teaserblue}{rgb}{0.274, 0.694, 1}
\definecolor{teaserorange}{rgb}{0.913, 0.443, 0.196}
\definecolor{teaserred}{rgb}{1,0,0}

\definecolor{kellygreen}{rgb}{0.3, 0.73, 0.09}

\newcolumntype{g}{>{\columncolor{CuGray}}c}
\newcolumntype{z}{>{\columncolor{CuGray}}l}

\renewcommand{\paragraph}[1]{\noindent\textbf{#1{.}}\,}

\definecolor{steelblue}{rgb}{0.27, 0.51, 0.7}

\usepackage{xspace}

\makeatletter
\def\@fnsymbol#1{\ensuremath{\ifcase#1\or *\or \dagger\or \ddagger\or
   \mathsection\or \mathparagraph\or \|\or **\or \dagger\dagger
   \or \ddagger\ddagger \else\@ctrerr\fi}}
\makeatother

\def\onedot{.\@\xspace}
\def\eg{\emph{e.g}\onedot} 
\def\ie{\emph{i.e}\onedot}

\newcommand{\Sref}[1]{Sec.~\ref{#1}}

\newcommand{\Fref}[1]{Fig.~\ref{#1}}
\newcommand{\Tref}[1]{Table~\ref{#1}}

\newcommand{\be}{\begin{eqnarray}}
\newcommand{\ee}{\end{eqnarray}}
\newcommand{\bee}{\begin{eqnarray*}}
\newcommand{\eee}{\end{eqnarray*}}

\newcommand{\matrixb}{\left[ \begin{array}}
\newcommand{\matrixe}{\end{array} \right]}

\newcommand{\cmark}{\ding{51}}%
\definecolor{graygray}{HTML}{bbbbbb}
\newcommand{\graycross}{{\color{graygray}\ding{55}}} 
\usepackage{mathtools}
\DeclarePairedDelimiter{\norm}{\lVert}{\rVert}

\tcbset{
    mytheoremstyle/.style={
        enhanced,
        breakable,
        colback=cyan!2,
        colframe=cyan!25!black,
        boxrule=0.4pt,
        arc=2mm,
        left=2mm,
        right=2mm,
        top=1mm,
        bottom=1mm,
        fonttitle=\bfseries,
    }
}

\definecolor{ExtFG}{HTML}{7B2CBF}   
\definecolor{OrigFG}{HTML}{A44A3F}  

\newtheorem{proposition}{Proposition}
\newtheorem{lemma}{Lemma}
\newtheorem{definition}{Definition}

\usepackage[table]{xcolor}
\usepackage{makecell}
\usepackage{booktabs}

\definecolor{YesColor}{RGB}{235,248,235}
\definecolor{PartialColor}{RGB}{255,249,230}
\definecolor{NoColor}{RGB}{252,235,237}

\newcommand{\Yes}{\cellcolor{YesColor}Yes}
\newcommand{\Partial}{\cellcolor{PartialColor}Partial}
\newcommand{\No}{\cellcolor{NoColor}No}
\title{FacEDiT: Talking Head Video Editing \\via Facial Motion Infilling}

\author{
    Kim Sung-Bin$^{1}$ \quad Joohyun Chang$^{2}$ \quad David Harwath$^{3}$ \quad Tae-Hyun Oh$^{2}$\\ \\
    $^{1}$POSTECH \quad $^{2}$KAIST \quad $^{3}$The University of Texas at Austin\\
}
\usepackage{capt-of}

\begin{document}

\maketitle

\begin{abstract}
Editing a local segment of a talking head video without reshooting the scene remains challenging and underexplored. Local edits, such as insertion, deletion, or substitution, require synthesizing a replacement segment while leaving the rest of the video unchanged. This involves variable-duration local rewriting while preserving identity, unedited regions, and boundary continuity, which standard speech-driven generation and lip synchronization methods do not directly address. Moreover, direct supervision is impractical, as it would require paired videos of the same person and scene that differ only in a local spoken segment. We instead formulate talking head video editing as facial motion infilling, a self-supervised pretext task that recovers masked facial motion from speech and surrounding motion context using ordinary video--speech pairs. Our key insight is that local video editing can be simulated during training by masking a motion span and reconstructing it from speech and visible motion context. Building on this formulation, we introduce FacEDiT, a mask-controlled talking head model with a local temporal attention bias and temporal smoothness regularization to improve lip--speech alignment and transition continuity. We also introduce FacEDiTBench, the first benchmark for talking head video editing, covering diverse edit types and lengths with dedicated evaluation metrics. Extensive experiments show that FacEDiT produces accurate, speech-aligned edits with strong identity preservation and seamless boundary transitions. Beyond editing, the same facial motion infilling model extends to portrait animation and lip synchronization by simply changing the mask pattern, establishing FacEDiT as a unified framework for talking head video editing and generation. 
Code and benchmark data are available at \url{https://facedit.github.io/}.
\end{abstract}

\section{Introduction} \label{sec:intro}

Filmmakers and media producers often need to revise specific parts of recorded videos, for example, when a word is misspoken or the script changes after shooting. In the iconic scene from Titanic (1997), Rose says, \textit{``I'll never let go, Jack,''} but a director might later prefer \textit{``I'll never let go no matter what happens, Jack.''} Traditionally, such changes require reshooting the scene, which is costly and time-consuming. Recent progress in \emph{talking head video synthesis} offers a promising alternative by generating or retargeting facial motion from speech. However, most existing methods are designed for speech-driven portrait animation or lip synchronization, rather than transcript-level edits that modify only a local video segment while leaving the remainder unchanged.

\begin{figure}[t]
    \centering
    \includegraphics[width=1\linewidth]{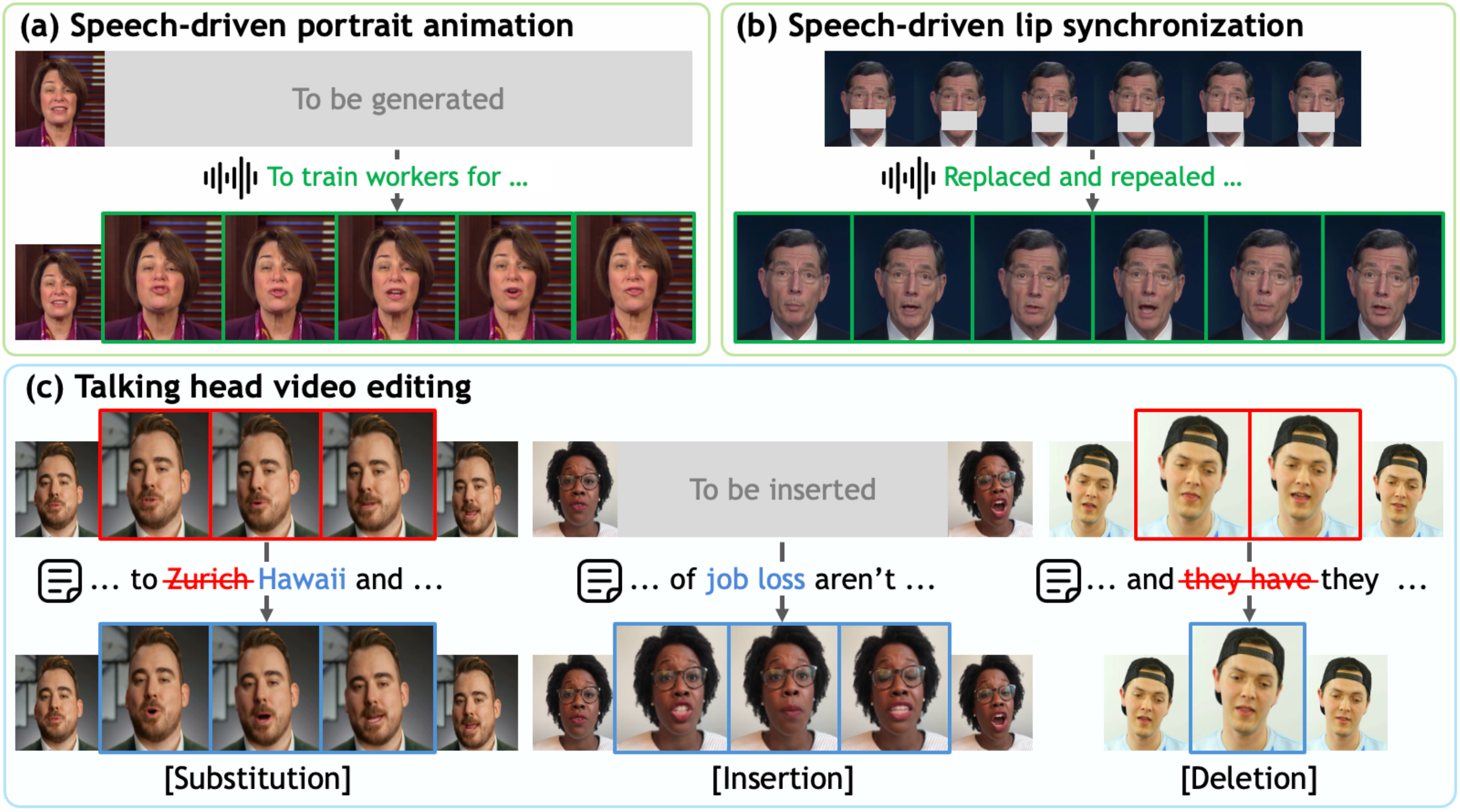}
    \vspace{-5mm}
    \caption{\textbf{Overview of FacEDiT.} FacEDiT performs mask-controlled facial motion infilling for local talking head video editing, regenerating only the edited span from speech and visible motion context. By changing the mask pattern, the same model supports portrait animation and lip synchronization.}
 \label{fig:teaser}
    
\end{figure}

Concretely, talking head video synthesis is commonly studied in two settings (\Fref{fig:teaser}):
\textit{(a) speech-driven portrait animation}, which animates a reference image from speech, and
\textit{(b) speech-driven lip synchronization}, which modifies mouth motion in an existing video to follow target speech.
In contrast, \textit{(c) talking head video editing} performs local edits, such as substitution, insertion, and deletion, within an existing video.
Although these tasks are closely related, talking head video editing poses distinct challenges: it must
(i) synthesize a local replacement segment whose duration may differ from the original,
(ii) preserve the original identity and all unedited regions, and
(iii) maintain seamless transitions at the boundaries between edited and unedited frames.

Several talking head video editing approaches have been proposed, but they often require per-speaker optimization~\cite{fried2019text,yao2021iterative} or degrade under long and complex edits~\cite{yang2023context}. 
A more fundamental obstacle is the lack of direct supervision, as paired videos of the same person and scene that differ only in a local spoken segment are impractical to obtain. As a result, speech-driven synthesis models are commonly repurposed as editing baselines, but their assumptions are mismatched. 
Portrait animation models~\cite{sadtalker,hallo,chen2025echomimic,hallo2,keyface,cui2025hallo3,echomimicv2,cha2025emotalkinggaussian} generate full sequences rather than preserving unedited regions, so stitching locally generated segments back into the original video often introduces boundary discontinuities. 
Meanwhile, lip synchronization models~\cite{kr2019towards,prajwal2020lip,zhou2021pose,mukhopadhyay2024diff2lip,li2024latentsync} mainly retarget mouth motion under a driving pose trajectory, making them insufficient for duration-changing edits when driving content is unavailable. 
These limitations suggest that talking head video editing should be treated as a primary problem rather than a minor variant of speech-driven generation or lip retargeting.

In this work, we address talking head video editing through the lens of \emph{facial motion infilling}. Given ordinary video--speech pairs, we construct a self-supervised pretext task by masking a temporal span of facial motion and training the model to recover the missing motion conditioned on speech and surrounding motion context. This converts editing from a supervised rewriting problem that requires paired edit data into a mask-conditioned reconstruction problem: the model learns to generate the edited region while using unmasked frames as context for identity preservation, motion continuity, and boundary consistency. Building on this formulation, we introduce \textbf{FacEDiT}, a mask-controlled talking head model that predicts compact facial motion latents from edited speech and visible motion context. To further improve lip--speech alignment and transition continuity, we incorporate a local temporal attention bias and temporal smoothness regularization.

Since no standard benchmark currently exists for talking head video editing, prior work lacks a common basis for evaluating whether an edit preserves identity and transitions smoothly across edit boundaries. To address this gap, we construct \textbf{FacEDiTBench}, a high-quality, human-verified dataset for this task, comprising 250 video and edited speech pairs. The dataset covers diverse edit types, edit lengths, and a wide range of identities. We also introduce dedicated metrics for identity preservation and boundary continuity to assess visual consistency and temporal smoothness in localized edits.

Extensive experiments show that FacEDiT enables accurate talking head video editing with precise lip synchronization, strong identity preservation, and seamless transitions, outperforming adapted baselines in quantitative and qualitative evaluations. Beyond editing, the same infilling strategy extends to related synthesis tasks by varying the mask pattern. Future-motion masking enables portrait animation, while lip-channel masking enables lip synchronization. This establishes FacEDiT as a unified framework for talking head editing and generation. Our contributions are summarized as:

\begin{itemize}
\item We cast talking head video editing as self-supervised facial motion infilling, learning variable-duration local edits from ordinary video--speech pairs without paired editing supervision.

\item We propose FacEDiT, a mask-controlled motion infilling model that combines temporal attention bias and temporal smoothness regularization for speech-aligned edits and seamless transitions.

\item We introduce FacEDiTBench, the first human-verified benchmark for talking head video editing, covering diverse edit types and lengths with dedicated identity and boundary-continuity metrics.

\item We demonstrate that the same facial motion infilling model generalizes to portrait animation and lip synchronization through simple changes to the mask pattern, establishing FacEDiT as a unified model for talking head editing and generation.
\end{itemize}

\section{Related work}\label{sec:related}
\paragraph{Talking head video editing}
Talking head video editing aims to modify only a local segment of an existing video according to transcript or speech edits, such as substitution, insertion, or deletion, while preserving the rest of the video.
Early methods retrieve viseme--phoneme segments followed by neural rendering~\cite{fried2019text}, perform iterative neural retargeting~\cite{yao2021iterative}, or predict edited motion with person-specific neural rendering~\cite{yang2023context,chen2025dfnerf}. These approaches often require target-speaker data or per-identity optimization, and their performance degrades for long or complex edits. Recent diffusion-based editing pipelines~\cite{han2024text} further explore local editing, but remain task-specific and do not provide a general formulation that also covers related synthesis settings. Despite its practicality, talking head video editing remains underexplored due to limited standardized benchmarks and public implementations. To address this gap, we introduce FacEDiTBench, a high-quality, human-verified benchmark with dedicated metrics for talking head video editing.

\begin{table}[t]
\caption{
\textbf{Comparison of talking head editing and synthesis settings.}
Duration-changing local edit denotes modifying a local segment in an existing video while allowing the edited segment length to change, such as insertion or deletion.
Unedited preservation indicates whether the method is designed to keep regions outside the edit unchanged.
Boundary continuity refers to modeling smooth transitions between edited and unedited regions.
Portrait animation and lip synchronization denote support for full-sequence speech-driven generation and mouth-motion retargeting, respectively.
}
\vspace{-1mm}
\label{tab:task_comparison}
\centering
\small
\resizebox{1\linewidth}{!}{%
\begin{tabular}{lccccc}
\toprule
\textbf{Method / Setting} &
\makecell{\textbf{Duration-changing}\\\textbf{local edit}} &
\makecell{\textbf{Unedited}\\\textbf{preservation}} &
\makecell{\textbf{Boundary}\\\textbf{continuity}} &
\makecell{\textbf{Portrait}\\\textbf{animation}} &
\makecell{\textbf{Lip}\\\textbf{sync}} \\
\midrule
Portrait animation~\cite{sadtalker,aniportrait,hallo,hallo2,keyface,vexpress,chen2025echomimic,echomimicv2,cui2025hallo3}
& \No & \No & \No & \Yes & \Partial \\

Lip synchronization~\cite{kr2019towards,prajwal2020lip,zhou2021pose,mukhopadhyay2024diff2lip,li2024latentsync,peng2025omnisync,ma2025sayanything}
& \Partial & \Partial & \Partial & \No & \Yes \\

\midrule
\textbf{FacEDiT}
& \Yes & \Yes & \Yes & \Yes & \Yes \\
\bottomrule
\end{tabular}%
}
\end{table}

\paragraph{Speech-driven synthesis as editing baselines}
Speech-driven talking head synthesis is commonly studied as portrait animation or lip synchronization. Portrait animation generates a full talking head video from a reference image and speech, using explicit motion intermediates such as 3DMM coefficients~\cite{sadtalker} or landmarks~\cite{aniportrait}, as well as recent diffusion and transformer-based generators~\cite{hallo,hallo2,keyface,vexpress,chen2025echomimic,echomimicv2,cui2025hallo3}. Lip synchronization retargets mouth motion in an existing video to match target speech, using GAN-based or audio-visual supervision~\cite{kr2019towards,prajwal2020lip}, pose-disentangled representations~\cite{zhou2021pose}, or diffusion-based inpainting and synchronization models~\cite{mukhopadhyay2024diff2lip,li2024latentsync,peng2025omnisync,ma2025sayanything}. 
These models provide strong baselines for talking head video editing, but their assumptions are structurally mismatched with variable-duration local editing: portrait animation does not preserve fixed unedited frames, and lip synchronization assumes an available driving pose and focuses mainly on the mouth region.

\paragraph{Facial motion infilling for unified editing and generation}
Recent talking head methods increasingly share compact motion representations, transformer backbones, and diffusion or flow-matching objectives, yet they remain tied to different task assumptions. 
As summarized in \Tref{tab:task_comparison}, existing settings do not jointly support duration-changing local edits, unedited-region preservation, boundary continuity, portrait animation, and lip synchronization. 
We instead formulate these settings as mask-controlled facial motion infilling, where temporal or channel-wise masks define the target task and the model predicts masked facial motion from speech and visible motion context. 
This formulation directly addresses the supervision gap in editing while enabling a single model to support both editing and generation without task-specific architectures.
\section{Method}\label{sec:method}
We propose FacEDiT, a mask-controlled talking head video editing model based on self-supervised facial motion infilling. Given speech and visible motion context, FacEDiT learns to reconstruct missing facial motion within a masked temporal span, matching the editing setting where only a local segment is regenerated while neighboring frames remain fixed. The following subsections describe the infilling formulation, training objective, inference pipeline, and interpretation of our formulation.

\subsection{Facial motion infilling formulation}\label{sec:formulation}
We formulate talking head video editing as \emph{facial motion infilling}. Given paired speech and video, we extract speech features $\mathbf{A}\in\mathbb{R}^{N\times D}$ and facial motion latents $\mathbf{F}=(\mathbf{f}_1,\dots,\mathbf{f}_T)\in\mathbb{R}^{T\times C}$, where $N$ and $T$ denote the numbers of speech and video frames, and $D$ and $C$ denote the speech and motion feature dimensions. We sample a binary spatiotemporal mask $\mathbf{M}\in\{0,1\}^{T\times C}$, where $\mathbf{M}_{t,j}=1$ indicates a motion entry to be predicted. Let $\mathbf{F}_{\mathbf{M}}=\mathbf{M}\odot\mathbf{F}$ denote the target motion entries and $\mathbf{F}_{\bar{\mathbf{M}}}=(1-\mathbf{M})\odot\mathbf{F}$ be the visible context. The goal is to learn the mask-conditioned distribution $p(\mathbf{F}_{\mathbf{M}} \mid \mathbf{F}_{\bar{\mathbf{M}}}, \mathbf{A})$ to infer missing facial motion from speech and visible motion context.

For talking head video editing, $\mathbf{M}$ covers the local span to be rewritten, so the model regenerates only the edited region while conditioning on surrounding unedited motion. The same formulation supports related synthesis tasks by changing the mask specification. For portrait animation, $\mathbf{M}$ masks all future motion after an observed initial prefix, allowing the model to complete the motion from speech. For lip synchronization, $\mathbf{M}$ masks only the lip-related channels while keeping pose and non-lip motion visible. Thus, a single model is controlled by the mask pattern at inference time.

\subsection{Training FacEDiT}\label{sec:training}
\begin{figure}[t]
    \centering
    \includegraphics[width=1\linewidth]{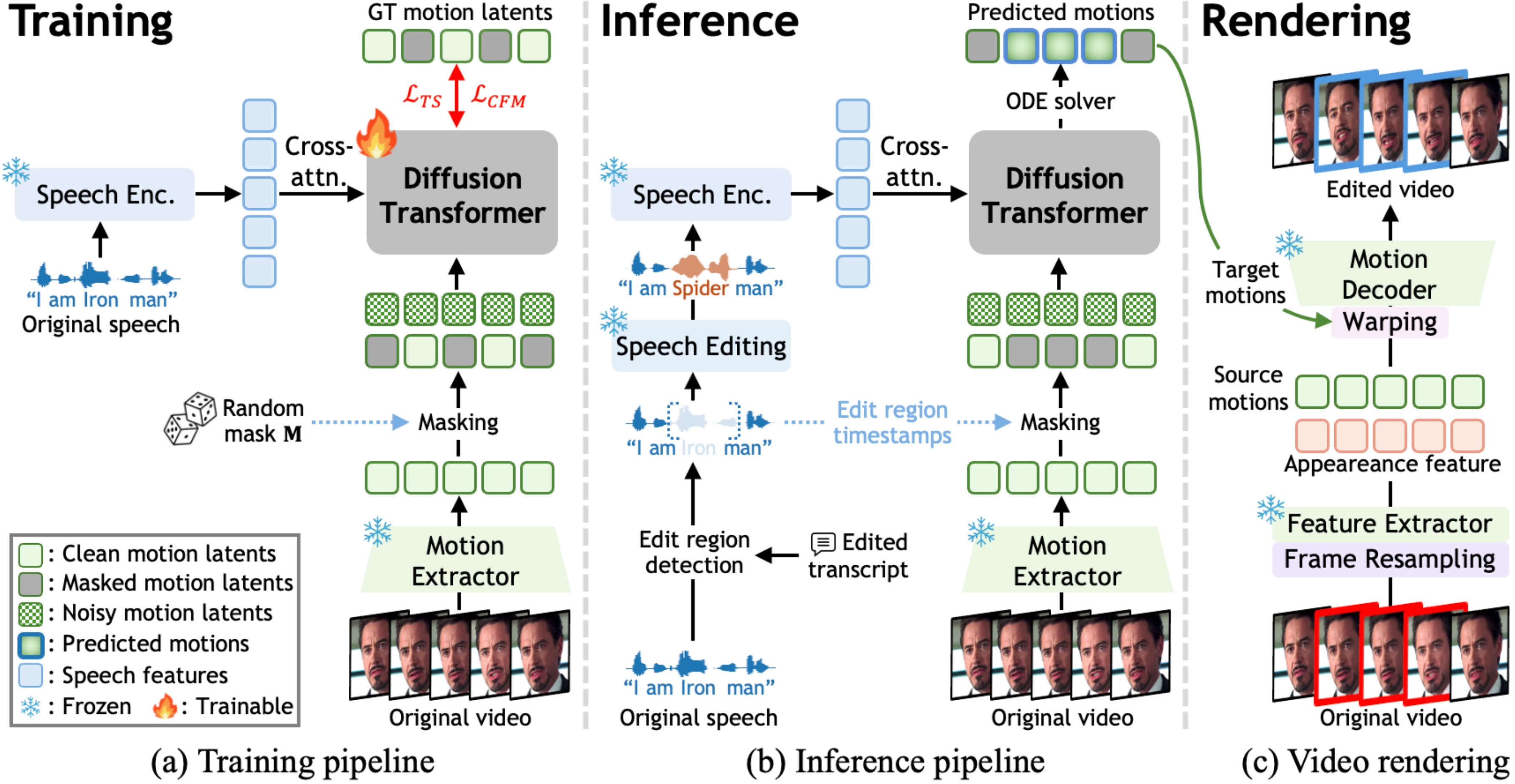}
    \vspace{-3mm}
    \caption{\textbf{Overview of FacEDiT}. (a) During training, FacEDiT learns facial motion infilling from paired speech and visible motion context using flow matching ($\mathcal{L}_{\text{CFM}}$) and temporal smoothness regularization ($\mathcal{L}_{\text{TS}}$). 
    (b) During inference, the edit span or task setting defines the motion mask, and speech guides the masked motion prediction. 
    (c) The predicted motion drives the source appearance features through the motion decoder to render the final video.} 
    \label{fig:pipeline}
\end{figure}

As illustrated in \Fref{fig:pipeline}~(a), we train FacEDiT to solve the facial motion infilling objective in \Sref{sec:formulation} using conditional flow matching (CFM)~\cite{lipman2023flow,liu2023flow}. Given ground-truth motion $\mathbf{F}_1$, Gaussian noise $\mathbf{F}_0$, and $t\sim\mathcal{U}[0,1]$, CFM defines the linear interpolation $\mathbf{F}_t=(1-t)\mathbf{F}_0+t\mathbf{F}_1$ with target velocity $\mathbf{F}_1-\mathbf{F}_0$. 
Given a sampled mask $\mathbf{M}$, the model is trained to reconstruct the target motion entries $\mathbf{M}\odot\mathbf{F}_1$ from noisy motion and the visible context $\mathbf{F}_{\text{cond}}=\mathbf{F}_{\bar{\mathbf{M}}}=(1-\mathbf{M})\odot\mathbf{F}_1$, conditioned on speech features $\mathbf{A}$.
During training, we randomly sample the temporal location and length of the masked span, while the channel-wise mask follows one of two predefined patterns. Full-motion masking hides all facial motion dimensions within the selected temporal span, whereas lip-only masking hides only the lip-related channels while preserving the remaining expression, head rotation, and translation components~\cite{kwon2025jam}. This mixed masking strategy exposes the model to the mask patterns used across editing and related synthesis tasks.

\paragraph{Model architecture}
We use the Diffusion Transformer (DiT)~\cite{peebles2023scalable} as the backbone of our motion infilling model. The noisy motion and visible motion are concatenated along the channel dimension and projected into the DiT latent space, where convolutional positional embeddings are added. For speech conditioning, a straightforward approach is to concatenate the speech features with the noisy and visible motion features at the input level, referred to as early fusion. However, we empirically find that injecting speech features through multi-head cross-attention in each DiT block achieves better synchronization between lip movements and speech, as well as improved overall performance (see \Tref{tab:ablation_all}). We employ scaled rotary positional embeddings (RoPE)~\cite{su2024roformer} to encode temporal ordering in both the self- and cross-attention layers.

\paragraph{Temporal multi-head attention bias}
To encourage seamless transitions and precise lip-speech alignment, we incorporate a local temporal attention bias. Specifically, we add a bias $\mathbf{B}$ to the query--key attention logits in both self- and cross-attention, where $\mathbf{B}(i,j)=0$ if $|i-j|<w$ and $-\infty$ otherwise. Given the projected query, key, and value features $\mathbf{Q}$, $\mathbf{K}$, and $\mathbf{V}$, attention is computed as $\mathrm{softmax}(\mathbf{Q}\mathbf{K}^{\top}/\sqrt{d_k}+\mathbf{B})\mathbf{V}$, where $d_k$ is the key dimension. This restricts each frame to attend to a local temporal window, encouraging the generated segment to connect smoothly with surrounding unedited motion and promoting local alignment between speech and motion features.

\paragraph{Learning objective}
Our main objective is the mask-conditional CFM loss, which trains the model to predict the velocity field conditioned on speech, visible motion context, and the mask:
\begin{equation}
\label{eq:cfm}
    \mathcal{L}_{\text{CFM}}
    =
    \mathbb{E}_{\mathbf{M}\sim q(\mathbf{M}),\,t,\,\mathbf{F}_0,\,\mathbf{F}_1,\,\mathbf{A}}
    \Big[
    \norm{
    \mathbf{M}\odot
    \left(
    v_\theta(\mathbf{F}_t,t,\mathbf{A},\mathbf{F}_{\text{cond}},\mathbf{M})
    -
    (\mathbf{F}_1-\mathbf{F}_0)
    \right)
    }_2^2
    \Big],
\end{equation}
where $\mathbf{F}_t=(1-t)\mathbf{F}_0+t\mathbf{F}_1$,
$q(\mathbf{M})$ is the training mask distribution, and $v_\theta$ is our DiT-based model.
To stabilize predicted motion dynamics and improve transition smoothness, we additionally apply a temporal smoothness loss
$\mathcal{L}_{\text{TS}}=\frac{1}{T-1}\sum_{k=2}^{T}\norm{\hat{\mathbf{F}}_1^k-\hat{\mathbf{F}}_1^{k-1}}_1$,
where the clean motion estimate is obtained from the predicted velocity as
$\hat{\mathbf{F}}_1=\mathbf{F}_t+(1-t)\,v_\theta(\mathbf{F}_t,t,\mathbf{A},\mathbf{F}_{\text{cond}},\mathbf{M})$.
The final objective is
$\mathcal{L}_{\text{total}}=\mathcal{L}_{\text{CFM}}+\lambda_{\text{TS}}\mathcal{L}_{\text{TS}}$,
where $\lambda_{\text{TS}}$ controls the smoothness weight.

\subsection{Inference}\label{sec:inference}
\paragraph{Talking head video editing}
\Fref{fig:pipeline}~(b) shows our inference pipeline for talking head video editing. Unlike training, the model conditions on edited speech, and the mask is defined by the edit span. Given an original video-speech pair and an edited transcript, we use word-level forced alignment to localize the segment to be modified, and synthesize the edited speech with a speech editing model. We then extract facial motion latents from the original video, mask the localized segment, and adjust its duration to match the edited speech. Finally, the noisy motion, visible motion context, and edited speech are fed into the model to generate the final facial motion sequence.

\paragraph{Speech-driven portrait animation and lip synchronization}
For speech-driven portrait animation, we use a future-motion mask that extends beyond the source motion to match the target speech duration. The model predicts the masked future motion with the available source motion and target speech features. For lip synchronization, we apply a lip-only channel mask while keeping head pose and expression dimensions unmasked; the remaining steps follow the editing procedure.

\paragraph{Video rendering and frame resampling}
After predicting facial motion, we use it as the driving signal to animate the video. As shown in \Fref{fig:pipeline}~(c), we extract appearance features and source motion from the original video, warp the source features using the predicted motion, and feed them into the frozen motion decoder to render the final edited frames while preserving identity. For duration-changing edits, the original frames must be resampled to match the edited speech length. To this end, we design a face-aware optical-flow-based resampling module that reduces artifacts from naive duplication, interpolation, or sparse sampling. Details are provided in Appendix~\ref{app:resampling}.

\subsection{Interpretation of facial motion infilling}
We provide a theoretical interpretation of FacEDiT as a mask-conditional generative model. 
The key observation is that different talking head synthesis tasks can be represented within a single model through different specifications of the mask $\mathbf{M}$.
\begin{proposition}[Task unification via motion infilling]
\label{thm:task_unification_main}
Talking head video editing, speech-driven portrait animation,
and speech-driven lip synchronization can all be expressed as facial motion infilling problems by choosing different masks \(\mathbf{M}\).
\end{proposition}
For editing, $\mathbf{M}$ masks a local temporal span to be rewritten; for portrait animation, it masks future motion after an initial prefix; and for lip synchronization, it masks lip-related motion channels while keeping pose and non-lip motion visible. 
Thus, the same conditional distribution $p(\mathbf{F}_{\mathbf{M}} \mid \mathbf{F}_{\bar{\mathbf{M}}}, \mathbf{A})$ can represent all three task settings by changing only the mask specification.

Beyond the formulation-level view in Proposition~\ref{thm:task_unification_main},
we show how random-mask training under conditional flow matching (CFM) enables a single model to amortize a family of mask-conditioned velocity fields and generalize across task masks at inference time.
\begin{proposition}[Per-mask optimality under coverage]
Let \(\mathcal{M}_{\mathrm{task}}\) be the set of masks required for tasks during inference. 
Under mild assumptions and sufficient model capacity, the optimal model obtained through random-mask training with the CFM objective in Eq.~\eqref{eq:cfm} recovers the Bayes-optimal conditional velocity field over the masked entries for each task mask \(\mathbf{M}^\star\in\mathcal{M}_{\mathrm{task}}\).
\end{proposition}
This result suggests that, if the training mask distribution sufficiently covers the masks used at inference time, random-mask training allows FacEDiT to generalize across diverse task-specific masking patterns within a single conditional generation framework.
More broadly, the unified formulation may extend beyond these three tasks,
\eg, to facial expression generation while keeping lip motion fixed.
Nevertheless, the proposition should be interpreted as an idealized
population-risk analysis rather than a guarantee for finite-capacity models.
Detailed assumptions, discussion of the finite-capacity setting, and proofs are
provided in Appendix~\ref{app:theory}.

\subsection{Implementation details}\label{sec:impl}

For facial motion representation and rendering, we adopt the frozen
LivePortrait pipeline~\cite{guo2024liveportrait}, which provides a compact
per-frame motion representation and renderer, while FacEDiT predicts
speech-conditioned masked motion sequences and their transitions to the visible
context. Specifically, we use the expression deformation, head rotation, and
translation components $[\delta,R,\tau]\in\mathbb{R}^{25\times3}$, flattened
into $\mathbf{f}\in\mathbb{R}^{75}$, corresponding to $C=75$ in
\Sref{sec:formulation}. This representation separates motion from appearance,
allowing FacEDiT to modify facial dynamics while preserving fixed appearance
features. Our model consists of 22 DiT layers with 16 attention heads and
hidden/feed-forward dimensions of 1024/2048. VoiceCraft~\cite{peng2024voicecraft}
is used for speech editing, while WavLM~\cite{chen2022wavlm} encodes speech into
$D$-dimensional features ($D{=}768$), which are projected into the DiT space.
The LivePortrait motion and appearance extractors and decoder, as well as WavLM,
are frozen; only FacEDiT's DiT and projection layer are trained.

For training, we use AdamW with a peak learning rate of $1\text{e-}4$, linearly warmed up for 20K steps and decayed thereafter. We train on two A6000 GPUs with a total batch size of 8 for up to 1M steps with early stopping. The temporal smoothness weight $\lambda_{\text{TS}}$ is set to 0.2. During inference, exponential moving average weights with a decay rate of 0.999 are used to stabilize predictions. Sampling uses the Euler ODE solver with timestep scaling based on the Sway strategy~\cite{chen2025f5}.

\section{FacEDiTBench: Benchmark for talking head video editing}\label{sec:dataset}
\paragraph{Dataset} To support thorough evaluation of talking head video editing, we construct FacEDiTBench, a first-of-its-kind dataset with 250 manually curated samples.
Each sample includes an original video, original speech, original transcript, edited transcript, and edited speech. 
Videos are sourced from 100 samples from HDTF~\cite{zhang2021flow}, 100 from Hallo3~\cite{cui2025hallo3}, and 50 from CelebV-Dub~\cite{sung2025voicecraft}, with the quality of both video and audio manually verified for each sample. 

To construct the edited pairs, we prompt GPT-4o~\cite{hurst2024gpt} to revise the
original transcript to generate grammatically correct edits, and use
VoiceCraft~\cite{peng2024voicecraft} with the original speech and edited transcript
to synthesize the corresponding edited speech.
Both the edited transcript and synthesized speech are manually verified for quality.
For each sample, we annotate the type of modification, the start and end timestamps of the edited segment, and the editing span length. Specifically, the modification types include insertion, deletion, and substitution, while the editing span length varies from short (1--3 words) to medium (4--6 words) and long (7--10 words). FacEDiTBench does not include paired ground-truth edited videos, as obtaining near-identical recordings that differ only in a local spoken segment is impractical.
Dataset statistics and an insertion example are shown in \Tref{tab:stat}.

\begin{table}[t]
    \centering
    \renewcommand{\arraystretch}{0.9}
    \caption{\textbf{Statistics and example of FacEDiTBench.}
    (a) FacEDiTBench includes three types of edits with span lengths ranging
    from short to long. (b) An example of transcript editing shows a phrase
    insertion while preserving the surrounding content.}
    \label{tab:stat}
    \vspace{-1mm}

    \begin{minipage}[t]{0.58\columnwidth}
        \centering
        \resizebox{0.98\linewidth}{!}{%
        \begin{tabular}{l|ccc|c}
        \toprule
        \diagbox{\textbf{Edit span}}{\textbf{Edit type}}
        & \textbf{Insertion}
        & \textbf{Substitution}
        & \textbf{Deletion}
        & \textbf{Total}\\
        \cmidrule{1-5}
        1--3 words (short)  & 9  & 20 & 5  & 34\\
        4--6 words (medium) & 43 & 70 & 11 & 124\\
        7--10 words (long)  & 29 & 59 & 4  & 92\\
        \bottomrule
        \end{tabular}}
    \end{minipage}
    \hfill
    \begin{minipage}[t]{0.41\columnwidth}
        \centering
        \resizebox{0.95\linewidth}{!}{%
        \begin{tabular}{c@{\quad}c}
        \toprule
        \textbf{Original transcript} & \textbf{Edited transcript}\\
        \cmidrule{1-2}
        \multirow{2}{*}{%
            \makecell{Father, it makes\\no sense to me.}}
        &
        \multirow{2}{*}{%
            \makecell{Father, \textbf{I can't do that,}\\
            it makes no sense to me.}}\\
        &\\
        \bottomrule
        \end{tabular}}
    \end{minipage}

    \vspace{1.0mm}

    \begin{minipage}[t]{0.58\columnwidth}
        \centering
        {\small (a) Statistics}
    \end{minipage}
    \hfill
    \begin{minipage}[t]{0.41\columnwidth}
        \centering
        {\small (b) Example of insertion edit type}
    \end{minipage}

    \vspace{-3mm}
\end{table}

\paragraph{Metrics} 
Along with the dataset, we introduce metrics for two key aspects of editing quality:
(1) boundary continuity, including photometric and motion continuity, and
(2) identity preservation. Since ground-truth edited videos are unavailable and
edited speech may correspond to multiple valid facial expressions, direct expression
accuracy cannot be evaluated. We therefore focus on identity preservation and boundary continuity.
\begin{itemize}
    \item \textbf{Photometric continuity (P$_{\text{continuity}}$ $\downarrow$)} measures pixel-level intensity changes at edit boundaries, indicating how naturally the color and lighting of the edited region blend with unedited areas.
\item \textbf{Motion continuity (M$_{\text{continuity}}$ $\downarrow$)} measures variations in optical flow magnitude at edit boundaries using RAFT~\cite{teed2020raft}. This metric captures the consistency of motion patterns across edited and unedited frames, reflecting temporal smoothness and stability in facial dynamics.
\item \textbf{Identity preservation (IDSIM $\uparrow$)} computes cosine similarity between identity features of original and edited videos using a face recognition model~\cite{deng2019arcface}.
\end{itemize}

\section{Experiments}\label{sec:exp}
\subsection{Experimental setup}\label{sec:exp_setup}
\paragraph{Dataset}
We train our model on a combination of video datasets totaling about 200 hours, including 130 hours from Hallo3~\cite{cui2025hallo3}, 60 hours from CelebV-Dub~\cite{sung2025voicecraft}, and 10 hours from HDTF~\cite{zhang2021flow}. Hallo3 and CelebV-Dub contain in-the-wild videos from diverse scenarios, such as movies and vlogs, while HDTF provides high-quality talking-head videos sourced from YouTube.

For talking head video editing evaluation, we use our proposed FacEDiTBench, which covers diverse editing scenarios and assumes that the speech has already been modified to match the edited transcript. We additionally evaluate speech-driven portrait animation and lip synchronization on the full HDTF test split, a standard benchmark for these tasks.

\paragraph{Competing methods}
We compare FacEDiT with nine portrait animation models~\cite{sadtalker,hallo,vexpress,aniportrait,chen2025echomimic,hallo2,keyface,cui2025hallo3,echomimicv2} and three lip synchronization models~\cite{li2024latentsync,prajwal2020lip,mukhopadhyay2024diff2lip} using their official checkpoints, primarily on talking head video editing, and also on portrait animation and lip synchronization. Note that these models are not designed for the editing task; therefore, we use them only as baselines.\footnote{Recent talking head video editing studies~\cite{chen2025dfnerf,han2024text}, as well as earlier related works, do not release dedicated open-source models for this setting. We therefore repurpose existing portrait animation or lip synchronization models as practical baselines.}

\paragraph{Objective metrics}
In addition to the metrics introduced in \Sref{sec:dataset}, we use standard
measures of synthesis quality. Lip-sync error (LSE-D) and confidence (LSE-C)
are computed with SyncNet~\cite{syncnet}, Fréchet Video Distance
(FVD)~\cite{fvd} measures distributional video realism using real video segments as the reference distribution, and LPIPS~\cite{zhang2018unreasonable}
measures perceptual similarity. For editing, metrics are computed only on the edited segment; for portrait animation and lip synchronization, they are computed on the full sequence. All metrics are
used across tasks except boundary continuity, which is specific to editing,
and LPIPS, which is used only for portrait animation.

\begin{table}[t]
    \caption{\textbf{Comparison on talking head video editing.} We adapt speech-driven portrait animation and lip synchronization models as editing baselines (source task $\rightarrow$ editing). $\dagger$ uses boundary frames from the original video as keyframes for smooth interpolation while editing, and $*$ conditions on the original head pose and predicts only facial expression. 
    }
    \vspace{-1mm}
    \label{tab:quan}
    \centering
    \renewcommand{\arraystretch}{0.9}
    \resizebox{0.78\linewidth}{!}{\begin{NiceTabular}{l@{\quad\,\,}c@{\quad\,\,}c@{\quad\,\,}c@{\quad\,\,}c@{\quad\,\,}c@{\quad\,\,}c}
    \toprule
    \RowStyle{\bfseries}
    Method & LSE-D $\downarrow$& LSE-C $\uparrow$ & IDSIM $\uparrow$ & FVD $\downarrow$ 
    
    & P$_\text{continuity}$ $\downarrow$& M$_\text{continuity}$ $\downarrow$
    \\
    \cmidrule{1-7}
    \multicolumn{7}{c}{\textbf{Speech-driven portrait animation baselines $\rightarrow$ Editing}}\\
    \cmidrule{1-7}
    V-Express~\cite{vexpress}& 7.719& 6.381& 0.845&201.392&40.668&11.470\\
    AniPortrait~\cite{aniportrait}& 10.601 &0.904&0.905&141.574&8.445&8.281\\
    EchoMimic~\cite{chen2025echomimic}&9.313&4.289 &  0.885& 158.631& 12.924& 17.892\\
    EchoMimicV2~\cite{echomimicv2}&  9.698 &  3.559&  0.866& 142.929& 12.357&16.513\\
    SadTalker~\cite{sadtalker}& 7.644& 6.202& 0.851&112.951&8.243&7.442\\
    Hallo~\cite{hallo}& 8.244 & 5.664& 0.909& 99.454&7.242& 6.429\\
    Hallo2~\cite{hallo2}& 8.205 &5.724 &0.918&95.256& 7.598&6.001\\
    Hallo3~\cite{cui2025hallo3}&  8.754 & 5.559& 0.880& 106.525& 7.651& 7.311\\
    KeyFace~\cite{keyface}&9.487 & 4.309 & 0.885& 110.619& 7.705& 7.220\\
    KeyFace$^\dagger$~\cite{keyface}&10.579&3.680&0.950&83.435&5.284&1.969\\
    \cmidrule{1-7}
    \multicolumn{7}{c}{\textbf{Speech-driven lip synchronization $\rightarrow$ Editing}}\\
    \cmidrule{1-7}
    Wav2Lip~\cite{prajwal2020lip}&\uline{7.202}&\textbf{7.212}&0.950&83.118&4.369&\uline{1.180}\\
    LatentSync~\cite{li2024latentsync}&7.639&5.820&\uline{0.964}&85.881&3.515&1.396\\
    Diff2Lip~\cite{mukhopadhyay2024diff2lip}&8.139&4.453&0.936&71.818&\uline{3.473}&1.686\\
    SadTalker$^*$~\cite{sadtalker}&7.852&5.847&0.852&98.893&7.713&6.803\\
    AniPortrait$^*$~\cite{aniportrait}&10.759&2.532&0.918&145.894&6.067&2.678\\
    \cmidrule{1-7}
    \cmidrule{1-7}
    \rowcolor{blue!5}
    Ours& \textbf{7.135}& \uline{6.670} & \textbf{0.966}&\textbf{61.930}&\textbf{2.420}&\textbf{0.800}\\
    \bottomrule
    \end{NiceTabular}}
\end{table}

\subsection{Results and analyses}
\begin{figure}[!t]
    \centering
    \includegraphics[width=1\linewidth]{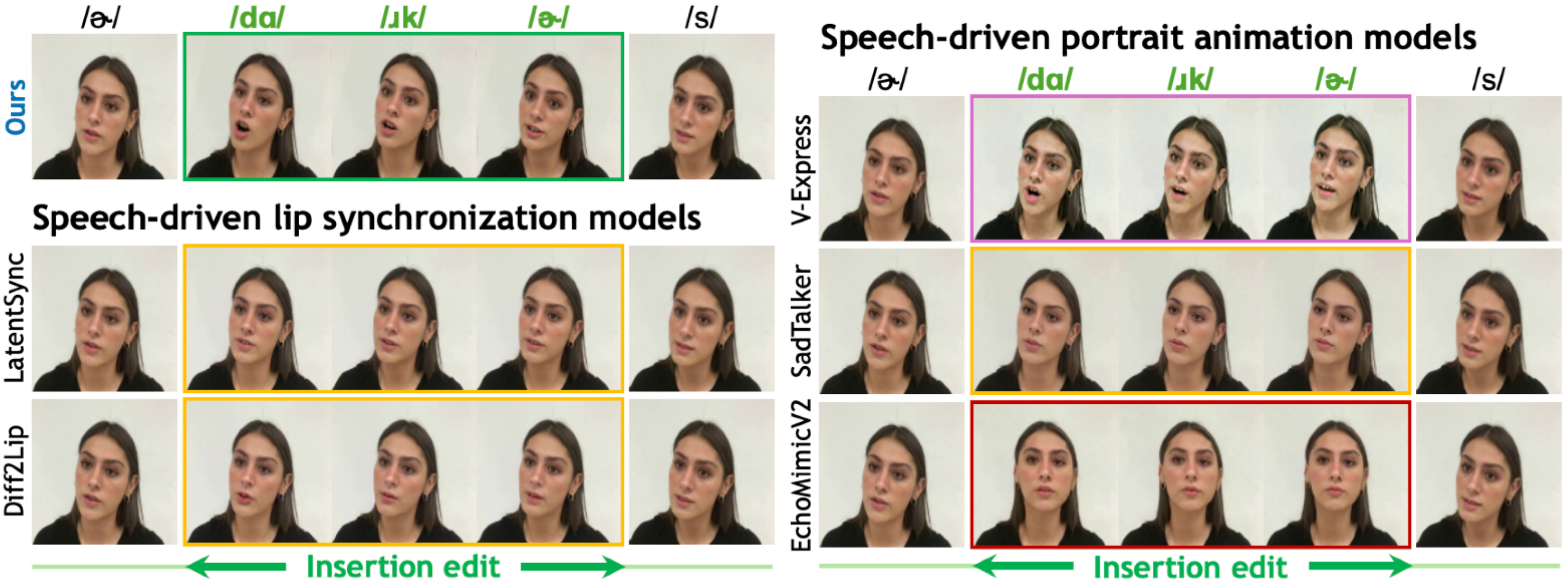}
    \vspace{-3mm}
    \caption{\textbf{Qualitative comparison.} The insertion edit adds the new word ``darker'' (\textcolor[HTML]{03B04F}{\textipa{/dA\*rk\textrhookschwa/}}) to the original sentence. The \textcolor[HTML]{03B04F}{green arrow} indicates the edited region, while the others are original frames.
    \textcolor[HTML]{0072B2}{Our model} achieves accurate lip-sync, preserves identity, and transitions smoothly to the original frames. In contrast, \textcolor[HTML]{D671CB}{V-Express} fails to preserve identity; \textcolor[HTML]{BA2116}{EchoMimicV2} shows a front-facing bias; \textcolor[HTML]{F8C32D}{SadTalker and lip synchronization models} produce mumbled motion; and most methods introduce discontinuities at boundary transitions (\eg, between \textcolor[HTML]{03B04F}{\textipa{/\textrhookschwa/}} and \textipa{/s/}). Full results are in Appendix~\ref{app:add_results}.}
    \label{fig:results}
\end{figure}

\paragraph{Comparison on talking head video editing}
Since publicly available talking head video editing models are limited, we repurpose speech-driven portrait animation and lip synchronization models as editing baselines. Portrait animation baselines synthesize only the edited segment and stitch it back into the original video, while lip synchronization baselines resample original frames to the edited speech length and apply lip editing with extracted head pose. We also include pose-conditioned SadTalker and AniPortrait variants ($*$), and a KeyFace extension ($\dagger$) using original boundary frames as keyframes. As shown in \Tref{tab:quan}, FacEDiT achieves the best boundary continuity and FVD while maintaining favorable lip-sync accuracy and identity preservation. Although KeyFace$^\dagger$ and lip synchronization models improve boundary continuity over portrait animation baselines, they still lag behind FacEDiT, indicating that such adaptations only partially address local video editing.

Figure~\ref{fig:results} shows a qualitative insertion example. Some baselines fail to preserve identity (\textcolor[HTML]{D671CB}{V-Express}), show a front-facing bias (\textcolor[HTML]{BA2116}{EchoMimicV2}), or produce mumbled motion (\textcolor[HTML]{F8C32D}{SadTalker and lip synchronization models}), and most show boundary discontinuities at the transition between \textcolor[HTML]{03B04F}{\textipa{/\textrhookschwa/}} and \textipa{/s/}. In contrast, our model preserves identity, produces smooth transitions, and generates lip motions consistent with the edited speech. Overall, these results show that local video editing requires dedicated modeling beyond adapted speech-driven synthesis baselines.

\begin{table}[!t]
    \centering
    \caption{\textbf{Robustness to edit duration.}  We evaluate 52 edits ranging
from 10 to approximately 40 seconds, grouped by edit duration, where $N$ denotes the number of edits in each group.}
    \label{tab:long_edit}

    \resizebox{0.78\linewidth}{!}{
    \begin{NiceTabular}{lccccccc}
    \toprule
    \RowStyle{\bfseries}
    Edit duration & $N$ & Avg. dur. &
    LSE-D $\downarrow$ & LSE-C $\uparrow$ &
    P$_{\text{continuity}}$ $\downarrow$ &
    M$_{\text{continuity}}$ $\downarrow$ &
    IDSIM $\uparrow$ \\
    \cmidrule{1-8}

    10--20 s
    & 18 & 15.89 s
    & 7.419 & 6.061
    & 1.259 & 0.360 & 0.9474 \\

    20--25 s
    & 13 & 21.64 s
    & 7.740 & 5.975
    & 1.204 & 0.312 & 0.9469 \\

    25--30 s
    & 9 & 26.93 s
    & 7.582 & 5.583
    & 1.772 & 0.388 & 0.9290 \\

    $\geq$30 s
    & 12 & 33.13 s
    & 7.673 & 5.178
    & 1.501 & 0.549 & 0.9200 \\

    \bottomrule
    \end{NiceTabular}}
\end{table}
\paragraph{Robustness to edit duration}
To evaluate FacEDiT beyond the edit lengths covered by FacEDiTBench, we
additionally evaluate 52 edits ranging from 10 to approximately 40 seconds,
grouped by edit duration (\Tref{tab:long_edit}).
As the edit duration increases, we observe no sharp performance degradation, although lip-sync confidence and identity similarity gradually decrease, while motion continuity tends to degrade for longer edits.
Based on these results, we conservatively regard approximately 30 seconds as
the currently validated practical range, rather than a strict maximum.

Several edits beyond approximately 40 seconds could not be reliably evaluated
because the upstream speech-editing model produced drifting speech.
We exclude these cases because the failure occurs in the upstream speech-editing stage before facial-motion generation, rather than in FacEDiT itself.
The relative robustness to long edits is consistent with our training setup,
where long masked spans are predicted jointly rather than autoregressively.

\begin{table}[tp]
    \centering
    \caption{\textbf{Ablation results on FacEDiTBench.} We evaluate the impact of different design choices by comparing various configurations. ``Early'' and ``CA'' indicate the speech feature integration strategy (early fusion vs.\ cross-attention). 
    (e) is the final model used in all experiments.}
    \label{tab:ablation_all}
    \resizebox{0.9\linewidth}{!}{\begin{NiceTabular}{lcccccccccc}
    \toprule
    \RowStyle{\bfseries}
    & Fusion & Speech Enc. &Biased Attn. & $\mathcal{L}_{\text{TS}}$ & LSE-D\(\,\downarrow\)& LSE-C\(\,\uparrow\) & IDSIM\(\,\uparrow\) & FVD\(\,\downarrow\) & P$_\text{continuity}$\(\,\downarrow\)& M$_\text{continuity}$\(\,\downarrow\)\\
    \cmidrule{1-11}
    (a) & Early & WavLM&\cmark & \graycross&7.292 & 6.581 & \uline{0.965}& 63.229& \uline{2.447}&0.825\\
    (b) & CA & WavLM&\cmark & \graycross& \uline{7.179} & \textbf{6.673} & \textbf{0.966}& \uline{61.955}&2.466&0.829\\
    (c) & CA & Wav2Vec&\cmark & \graycross&7.223 & 6.639 & \uline{0.965}& 62.171&2.448&\uline{0.821}\\
    (d) & CA & WavLM&\graycross & \graycross&7.301 & 6.655 & 0.955& 65.559& 2.816&1.097\\
    \rowcolor{blue!5}
    (e) & CA & WavLM&\cmark & \cmark&\textbf{7.135} & \uline{6.670} & \textbf{0.966}&\textbf{61.930} &\textbf{2.420}&\textbf{0.800}\\
    \bottomrule
    \end{NiceTabular}}
\end{table}

\paragraph{Ablation study}
We conduct ablations on FacEDiTBench to validate our design choices (\Tref{tab:ablation_all}). Comparing (b) with (a) and (c), cross-attention speech conditioning and WavLM~\cite{chen2022wavlm} improve lip synchronization over early fusion and Wav2Vec~\cite{baevski2020wav2vec}, confirming the benefit of stronger speech conditioning. Removing the attention bias ((b) vs.\ (d)) noticeably degrades overall performance, especially lip synchronization and boundary continuity (P$_{\text{continuity}}$, M$_{\text{continuity}}$), highlighting its importance for speech alignment and smooth transitions. Adding the temporal smoothness loss $\mathcal{L}_{\text{TS}}$ in (e) further improves boundary continuity and reduces the RMS velocity and acceleration of predicted motion by 5.6\% and 13.2\%, respectively, further supporting its role in stabilizing motion dynamics (Appendix~\ref{app:add_results}). Based on these results, we use (e) as the final model configuration.

\paragraph{Generalization to portrait animation and lip synchronization}
By changing mask patterns over facial motion latents, FacEDiT generalizes to portrait animation and lip synchronization. For portrait animation, we condition on the first frame and animate it with target speech; for lip synchronization, we follow the cross-generation setup~\cite{mukhopadhyay2024diff2lip}, where a driving video from the same subject provides identity and head pose while target speech drives lip motion. As shown in \Tref{tab:talking_combined}, FacEDiT achieves favorable performance on both tasks: it improves lip-sync accuracy and realism in portrait animation, with lower LSE-D and FVD, and remains competitive with specialized lip synchronization methods. These results demonstrate that motion infilling not only excels at editing but also transfers effectively to related tasks within a unified framework.

\begin{table}[!t]
\centering
\caption{\textbf{Results on cross-task generalization.}
Our unified model generalizes to speech-driven portrait animation and lip synchronization, achieving favorable performance.
}
\label{tab:talking_combined}

\begin{minipage}[c]{0.48\columnwidth}
\centering
\renewcommand{\arraystretch}{0.9}
\resizebox{1.1\linewidth}{!}{%
\begin{NiceTabular}{@{}l *{5}{c} @{}}
\toprule
\RowStyle{\bfseries}
Method
& \mbox{LSE-D\(\,\downarrow\)}
& \mbox{LSE-C\(\,\uparrow\)}
& \mbox{IDSIM\(\,\uparrow\)}
& \mbox{FVD\(\,\downarrow\)}
& \mbox{LPIPS\(\,\downarrow\)}\\
\midrule
V-Express~\cite{vexpress}&\uline{7.470}& \uline{7.842}&0.906 &65.158 & 0.341\\
AniPortrait~\cite{aniportrait}& 11.168& 2.866& 0.914&102.414 &0.310\\
EchoMimic~\cite{chen2025echomimic}& 9.085&5.555 & \uline{0.929}&89.129 &0.320\\
EchoMimicV2~\cite{echomimicv2}& 9.966& 4.051& 0.890& 131.603&0.322\\
SadTalker~\cite{sadtalker}& 7.557& 7.421&0.848 &81.572 & 0.309\\
Hallo~\cite{hallo}&7.824 &7.080 &0.925 &42.825 &0.283\\
Hallo2~\cite{hallo2}& 7.815& 7.038& 0.927& 43.130&\uline{0.281}\\
Hallo3~\cite{cui2025hallo3}& 8.583 & 6.717& 0.900& \uline{38.643}&0.315\\
KeyFace~\cite{keyface}& 8.905& 5.795& 0.926& 49.043&\textbf{0.267}\\
\midrule
\rowcolor{blue!5}
Ours& \textbf{6.950}& \textbf{7.960}& \textbf{0.930}&\textbf{31.662} &0.289\\
\bottomrule
\end{NiceTabular}}
\vspace{0.1mm}
\\{\small (a) Portrait animation}
\label{tab:talking}
\end{minipage}
\hfill
\begin{minipage}[c]{0.48\columnwidth}
\centering
\renewcommand{\arraystretch}{0.9}
\resizebox{0.9\linewidth}{!}{%
\begin{NiceTabular}{@{}l *{4}{c} @{}}
\toprule
\RowStyle{\bfseries}
Method
& \mbox{LSE-D\(\,\downarrow\)}
& \mbox{LSE-C\(\,\uparrow\)}
& \mbox{IDSIM\(\,\uparrow\)}
& \mbox{FVD\(\,\downarrow\)}\\
\midrule
Wav2Lip~\cite{prajwal2020lip} & 7.433 & \uline{7.635} & 0.931 & 42.843\\
Diff2Lip~\cite{mukhopadhyay2024diff2lip} & \textbf{6.910} & 7.475 & 0.913 & 29.580\\
LatentSync~\cite{li2024latentsync} & 7.177 & 7.533 & \uline{0.933} & \textbf{23.044}\\
\midrule
\rowcolor{blue!5}
Ours& \uline{7.132}& \textbf{7.661}& \textbf{0.935}&\uline{29.381}\\
\bottomrule
\end{NiceTabular}}
\vspace{1.5mm}
\\{\small (b) Lip synchronization}
\label{tab:lipsync}
\end{minipage}
\end{table}

\begin{table}[!t]
    \centering
    \caption{\textbf{Human study comparison.}
    We report the average rank ($\pm$ standard deviation) assigned by participants (lower is better)
    for talking head video editing and the corresponding generation task.
    Boundary continuity is evaluated for editing only.
    Ranks are computed independently within each comparison pool and are
    therefore not directly comparable across panels.
    }
    \label{tab:human}

    \resizebox{0.95\linewidth}{!}{%
    \begin{NiceTabular}{llcccccc}
    \CodeBefore
        \columncolor{blue!5}{3}
    \Body

    \toprule
    \textbf{Task} && \textbf{Ours}
    & \textbf{SadTalker}~\cite{sadtalker}
    & \textbf{Hallo}~\cite{hallo}
    & \textbf{Hallo2}~\cite{hallo2}
    & \textbf{Hallo3}~\cite{cui2025hallo3}
    & \textbf{KeyFace}~\cite{keyface}\\

    \midrule

    \multirow{2}{*}{\textbf{Editing}}
    & Overall $\downarrow$
    & \textbf{1.57}{\scriptsize$\pm$1.40}
    & 3.93{\scriptsize$\pm$1.38}
    & 3.58{\scriptsize$\pm$1.29}
    & 3.90{\scriptsize$\pm$1.40}
    & \uline{3.32}{\scriptsize$\pm$1.51}
    & 4.70{\scriptsize$\pm$1.51}\\

    & Boundary $\downarrow$
    & \textbf{1.39}{\scriptsize$\pm$1.25}
    & 4.19{\scriptsize$\pm$1.29}
    & 3.35{\scriptsize$\pm$1.07}
    & 3.90{\scriptsize$\pm$1.34}
    & \uline{3.23}{\scriptsize$\pm$1.49}
    & 4.93{\scriptsize$\pm$1.39}\\

    \midrule

    \textbf{Portrait animation}
    & Overall $\downarrow$
    & \textbf{2.32}{\scriptsize$\pm$1.60}
    & 4.83{\scriptsize$\pm$1.50}
    & 3.04{\scriptsize$\pm$1.35}
    & \uline{2.91}{\scriptsize$\pm$1.41}
    & 3.68{\scriptsize$\pm$1.59}
    & 4.23{\scriptsize$\pm$1.47}\\

    \bottomrule
    \end{NiceTabular}%
    }

    \vspace{1mm}

    {\small (a) Comparison with speech-driven portrait animation models}

    \vspace{1.7mm}

    \resizebox{0.8\linewidth}{!}{%
    \begin{NiceTabular}{llcccc}
    \CodeBefore
        \columncolor{blue!5}{3}
    \Body

    \toprule
    \textbf{Task} && \textbf{Ours}
    & \textbf{LatentSync}~\cite{li2024latentsync}
    & \textbf{Diff2Lip}~\cite{mukhopadhyay2024diff2lip}
    & \textbf{Wav2Lip}~\cite{prajwal2020lip}\\

    \midrule

    \multirow{2}{*}{\textbf{Editing}}
    & Overall $\downarrow$
    & \textbf{1.22}{\scriptsize$\pm$0.71}
    & 3.23{\scriptsize$\pm$0.87}
    & 2.83{\scriptsize$\pm$0.80}
    & \uline{2.72}{\scriptsize$\pm$0.88}\\

    & Boundary $\downarrow$
    & \textbf{1.23}{\scriptsize$\pm$0.67}
    & 3.19{\scriptsize$\pm$0.92}
    & \uline{2.74}{\scriptsize$\pm$0.77}
    & 2.83{\scriptsize$\pm$0.95}\\

    \midrule

    \textbf{Lip synchronization}
    & Overall $\downarrow$
    & \textbf{1.31}{\scriptsize$\pm$0.74}
    & 2.77{\scriptsize$\pm$0.93}
    & 3.20{\scriptsize$\pm$0.89}
    & \uline{2.71}{\scriptsize$\pm$0.90}\\

    \bottomrule
    \end{NiceTabular}%
    }

    \vspace{1mm}

    {\small (b) Comparison with speech-driven lip synchronization models}

\end{table}

\paragraph{Human evaluation}
We further conduct a human study to assess perceptual quality across editing
and generation tasks.
For each comparison, 30 participants evaluate 20 samples by ranking the
synthesized videos according to overall quality, including lip synchronization
and motion naturalness.
For talking head video editing, participants additionally evaluate boundary
continuity between the edited and unedited regions.
We conduct separate comparisons against portrait animation and lip
synchronization models, and ranks are computed independently within each
comparison pool. As shown in \Tref{tab:human}, FacEDiT is consistently preferred for talking
head video editing in both overall quality and boundary continuity, with a
clear margin over the competing methods.
It also achieves favorable rankings on portrait animation and lip
synchronization, showing that its favorable objective performance is also reflected in
perceptual preference across tasks.

\section{Conclusion and discussion}\label{sec:conclusion}
In this work, we presented FacEDiT, a mask-controlled framework that formulates talking head video editing as facial motion infilling. By masking the local edit span and reconstructing the missing facial motion from speech and surrounding motion context, FacEDiT enables variable-duration local edits while preserving identity, unedited regions, and boundary continuity. Beyond editing, the same infilling formulation extends to portrait animation and lip synchronization through future-motion and lip-channel masks, enabling a unified model for talking head editing and generation. We also introduced FacEDiTBench, the first dedicated benchmark for talking head video editing, with diverse edit types and dedicated metrics. Extensive experiments demonstrate accurate speech-aligned edits, strong visual consistency, and effective cross-task generalization, highlighting the potential of facial motion infilling for practical talking head video editing and synthesis.

While our method achieves strong performance in talking head video synthesis, it also has limitations. Since FacEDiT relies on the frozen LivePortrait motion extractor and renderer, it inherits their failure modes under extreme profile views, severe occlusions, and rapid head turns, which may lead to unreliable motion features or rendering artifacts. Occasional visual jumps and stale backgrounds may also remain due to the frozen pipeline, even when the predicted facial motion is plausible. Our model focuses on facial motion and therefore does not drive upper-body gestures from speech. In addition, emotional context is learned implicitly rather than modeled explicitly. Improving the motion extraction and rendering pipeline, extending the framework with modules for full-body motion, and incorporating explicit emotional cues could further improve realism.

\begin{ack}
\paragraph{Funding} This work was supported by the National Research Foundation of Korea (NRF) funded by the Korean government (MSIT) in 2026 (Grant No. RS-2024-00451947, Project Title: Interaction Framework for Connecting Users to Generative Experiences via Time-Space Expanded Multimedia: Centering on Social Communication Disorders (33\%); Grant No. RS-2024-00453301, Project Title: MASC: Convergence of Multimodal AI and Semantic Communications for Hyper-Connected Hyper-Intelligent Public Safety Systems (27\%)), the Institute of Information \& Communications Technology Planning \& Evaluation (IITP) grant funded by the Korean government (MSIT) (No. RS-2025-25441313, Professional AI Talent Development Program for Multimodal AI Agents (30\%); No. RS-2026-25518317, Development of AI memory mechanism that reflects human cognitive principles (10\%)), the InnoCORE program of the Ministry of Science and ICT (26-InnoCORE-01; N10250156), and the Artificial Intelligence Industrial Convergence Cluster Development Project funded by the Ministry of Science and ICT (MSIT, Korea) \& Gwangju Metropolitan City.

\paragraph{Competing interests} The authors declare no competing interests.
\end{ack}

\bibliographystyle{unsrt}
{\small
\bibliography{reference}
}

\newpage
\appendix
\pagebreak
\clearpage
\renewcommand{\thefigure}{S\arabic{figure}}
\renewcommand{\thetable}{S\arabic{table}}
\setcounter{figure}{0} 
\setcounter{table}{0} 
\setcounter{proposition}{0}

\startcontents[appendix]
\section*{Table of Contents}
{
    \hypersetup{linkcolor=black}
    \printcontents[appendix]{}{1}{}
}

\section{Details on FacEDiTBench}\label{app:faceditbench}
We introduce FacEDiTBench, a dedicated dataset and evaluation metrics for the talking head video editing task. Below, we describe the dataset construction pipeline and the evaluation metrics proposed for this task.

\subsection{Dataset}
As introduced in the main paper, FacEDiTBench contains diverse, high-quality video–speech pairs designed to evaluate talking head video editing.
Each sample includes the original video and its corresponding speech, along with an edited transcript and edited speech for direct evaluation.
The edited speech covers multiple edit types, including substitution, insertion, and deletion, with edit lengths ranging from one to ten words. \Tref{tab:dataset} shows examples of transcripts for each edit type, which are later used to generate the edited speech.
Human verification is incorporated at every stage to ensure dataset quality and accuracy.

\paragraph{Collecting video sources}
We construct our dataset using videos from HDTF~\cite{zhang2021flow}, CelebV-Dub~\cite{sung2025voicecraft}, and Hallo3~\cite{cui2025hallo3}.
CelebV-Dub and Hallo3 provide expressive, in-the-wild content across diverse sources, such as vlogs, dramas, and influencer videos, while HDTF offers high-quality interview videos from YouTube.
Combining these datasets allows FacEDiTBench to cover a broad range of speaking styles, environments, and visual conditions.

\paragraph{Detecting English and labeling pseudo-transcripts}
Since HDTF and Hallo3 do not include transcripts, we use WhisperX~\cite{bain2023whisperx} to automatically detect the language and generate pseudo-transcripts.
We retain only English-speaking videos and manually verify all generated transcripts for accuracy.

\begin{table}[tp]
\centering
\caption{\textbf{Examples of edit types in FacEDiTBench.} Each example shows the edit type and the number of edited words, along with the corresponding original and edited transcripts. The edited transcripts are used to generate edited speech for the dataset using VoiceCraft~\cite{peng2024voicecraft}.}
\resizebox{1\linewidth}{!}{
  \begin{tabular}{l@{\quad \quad}c@{\quad \quad}c} 
    \toprule
    Edit type&Original transcript & Edited transcript \\
    \midrule
    \multirow{2}{*}{\makecell[l]{Substitution\\(5 words)}} & \multirow{2}{*}{\makecell{However, if \textbf{that naturally} \\ \textbf{suits you}, then do}}& \multirow{2}{*}{\makecell{However, if \textbf{your heart} \\ \textbf{tells you to}, then do}}\\
    & &\\
    \midrule
    \multirow{3}{*}{\makecell[l]{Insertion\\(6 words)}} & \multirow{3}{*}{\makecell{And I forget the story behind it, \\but it is really cool to see}}& \multirow{3}{*}{\makecell{And I forget the story behind it, \\\textbf{about how the festival first began}, \\but it is really cool to see}}\\
    & & \\
    &&\\
    \midrule
    \multirow{2}{*}{\makecell[l]{Deletion\\(6 words)}} & \multirow{2}{*}{\makecell{So we are not looking at \textbf{a murder,}\\ \textbf{we are looking at} negligence}}& \multirow{2}{*}{So we are not looking at negligence}\\
     & & \\
    \bottomrule
  \end{tabular}
}
\label{tab:dataset}
\end{table}

\paragraph{Generating edited transcript}
For each verified transcript, we use GPT-4o~\cite{hurst2024gpt} to generate diverse edited variants. We prompt GPT-4o with the original transcript to create multiple candidates covering different edit types (substitution, deletion, and insertion) and varying edit lengths (1–10 words). We then manually verify the edited transcripts and label each sample with its corresponding edit type and length. After this stage, each sample contains the original video with the corresponding speech, and the original and edited transcripts. 

\paragraph{Generating edited speech}
Using the original speech, original transcript, and edited transcript, we synthesize edited speech with VoiceCraft~\cite{peng2024voicecraft}.
As VoiceCraft’s outputs vary with hyperparameters (\eg, duration or silence length), multiple samples are generated per edit, and we manually select the most natural and well-aligned version.

\paragraph{Human verification}
Finally, all samples undergo a rigorous human verification process to ensure quality. Any samples with low-quality edited speech or noticeable artifacts are discarded, leaving only high-fidelity pairs. The resulting dataset contains 250 verified samples, covering diverse video sources, speaking styles, and editing scenarios.

\subsection{Metrics}
Along with the dataset, we introduce three evaluation metrics for the talking head video editing task. Specifically, we propose \emph{photometric continuity} and \emph{motion continuity} to assess boundary continuity between the edited and unedited regions, measuring whether the transition is smooth in terms of pixel appearance and motion dynamics. We additionally propose an \emph{identity preservation} metric to evaluate whether the synthesized segment maintains the original identity.

\begin{figure}[!t]
    \centering
    \includegraphics[width=1\linewidth]{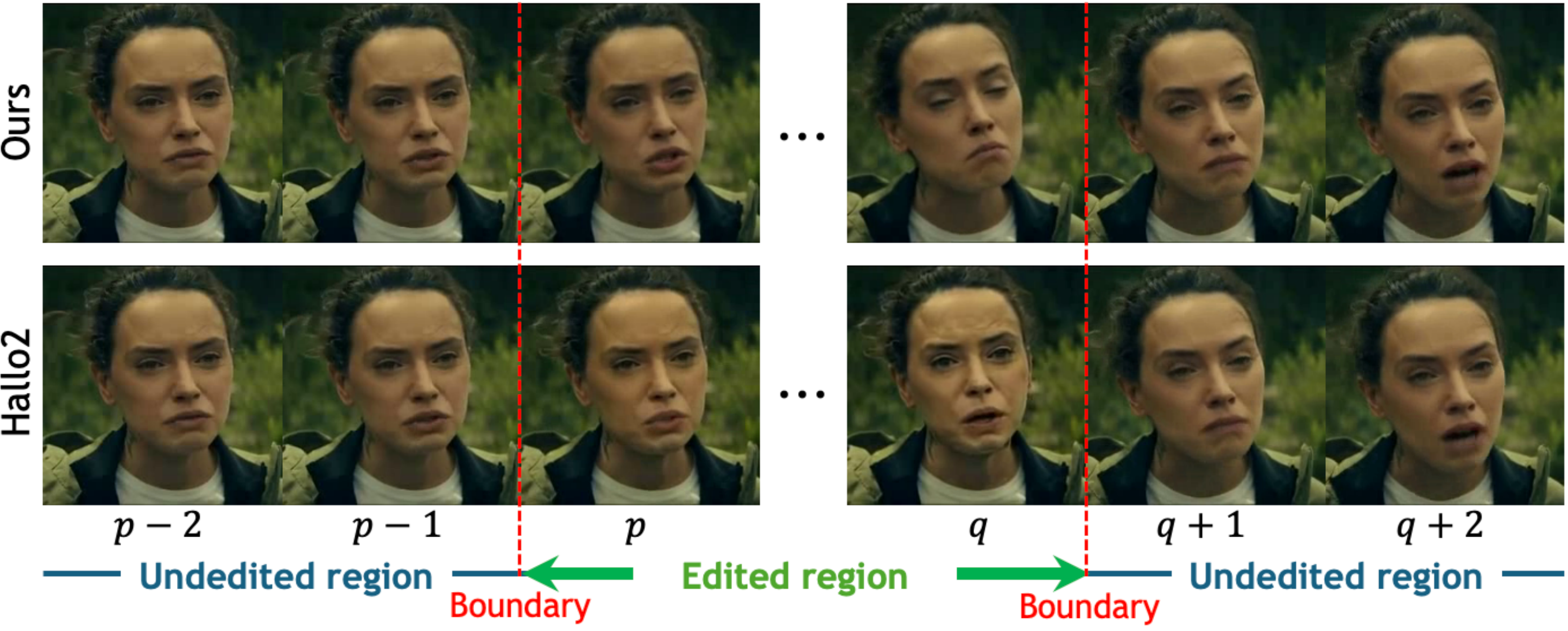}
    \caption{\textbf{Evaluation setup for boundary continuity.} Frames 1 to $(p-1)$ and $(q+1)$ to the last frame are treated as unedited, while frames $p$ through $q$ constitute the edited segment. Based on these segments, we compute the boundary continuity metric at the transition frames. Our model produces smooth transitions at these boundaries, whereas the comparison model (Hallo2) exhibits noticeable discontinuities, particularly between frames $q$ and $q+1$.}
    \label{fig:boundary}
\end{figure}

Given an edited video, we evaluate how smoothly the edited region blends with the surrounding unedited frames. Using the example in \Fref{fig:boundary}, let the first to $(p\!-\!1)$-th frames and the $(q\!+\!1)$-th to the last frames be unedited, while frames $p$ to $q$ constitute the edited segment. For boundary continuity, we measure how naturally the video transitions across the boundaries at $(p\!-\!1)\!\rightarrow\! p$ and $q\!\rightarrow\!(q\!+\!1)$. 

We illustrate the formulation using the right boundary as an example, although in practice boundary continuity is measured across all boundaries. We consider three consecutive frames $q$, $q\!+\!1$, and $q\!+\!2$, denoted by RGB images $I_q, I_{q+1}, I_{q+2} \in \mathbb{R}^{H\times W\times 3}$ with values in $[0,255]$. The edit occurs between $I_q$ and $I_{q+1}$, while $I_{q+2}$ is an unedited frame serving as a natural reference. Using this setup, we define two continuity metrics, photometric continuity and motion continuity, which together assess whether the edited boundary introduces unnatural temporal or motion discontinuities.

\paragraph{Photometric continuity}
This metric measures the second-order temporal derivative of pixel intensities across the boundary, capturing the photometric inconsistency introduced by the edited frame. Natural motion typically yields smooth temporal intensity trajectories, whereas abrupt or artificial edits lead to large photometric acceleration. For each pixel and color channel, we compute
\[
    \text{P}_{\text{continuity}} = \text{mean}\left( \left| I_{q+2} - 2I_{q+1} + I_q \right| \right),
\]
which corresponds to the discrete second derivative across the sequence. High values indicate that the edited frame deviates from the natural photometric trajectory, whereas low values indicate temporally coherent appearance and smooth transition across the boundary.

\paragraph{Motion continuity}
While photometric continuity captures intensity-level smoothness, motion continuity evaluates temporal consistency in the motion field using optical flow. Specifically, we examine whether the edited frame disrupts the expected symmetry between forward and backward motion around the boundary by measuring the change in optical-flow speed. We compute dense optical flow from the boundary frame to its neighboring frames as:
\begin{align*}
    \text{Forward flow}&: \; v_{q+1 \rightarrow q+2} \quad &&\text{from } I_{q+1} \text{ to } I_{q+2},\\
    \text{Backward flow}&: \; v_{q+1 \rightarrow q} \quad &&\text{from } I_{q+1} \text{ to } I_q.
\end{align*}
We then compute their magnitudes:
\[
    s_f = \lVert v_{q+1 \to q+2} \rVert_2,\qquad
    s_b = \lVert v_{q+1 \to q} \rVert_2.
\]
Temporal smoothness implies that the motion entering and exiting $I_{q+1}$ should have similar magnitudes. Thus, we define the motion continuity as:
\[
    \text{M}_{\text{continuity}} = \text{mean}\left( | s_f - s_b | \right).
\]
Only flow vectors valid in both directions are considered. A low $\text{M}_{\text{continuity}}$ indicates motion that transitions smoothly across the edited boundary, whereas a high value suggests unnatural acceleration or deceleration caused by the edit.

Photometric continuity evaluates temporal intensity smoothness, while motion continuity assesses motion-field consistency across the boundary. Together, they provide a comprehensive measure of whether the edited region blends seamlessly with the surrounding unedited content.

\paragraph{Identity preservation}
Identity preservation measures how well the synthesized segment maintains the original identity. We compute cosine similarity between the averaged identity features of the original and generated videos using a face recognition model~\cite{deng2019arcface}. Given an original video segment of $K$ frames $V_{1:K}$ and an edited video segment of $T$ frames $\hat{V}_{1:T}$, we extract facial features
\[
    f_{1:K} = \text{FaceEnc}(V_{1:K}), \qquad
    \hat{f}_{1:T} = \text{FaceEnc}(\hat{V}_{1:T}),
\]
where $\text{FaceEnc}(\cdot)$ denotes the face recognition model. We then compute the average feature vectors
\[
    f' = \text{mean}(f_{1:K}), \qquad
    \hat{f}' = \text{mean}(\hat{f}_{1:T}),
\]
and define the identity similarity as
\[
    \text{IDSIM} = \text{cosine\_similarity}(f', \hat{f}'),
\]
which serves as the identity preservation metric.

\section{Additional results and analysis}\label{app:add_results}
We provide additional experimental results and analyses that are not included in the main paper due to space limitations.

\paragraph{Additional comparison with portrait animation models}
In the main paper, we repurpose portrait animation baselines for talking head video editing by synthesizing only the locally edited region and stitching it back, which can introduce boundary artifacts. Here, we instead repurpose these baselines to generate the \emph{entire} video from the edited speech. While full-sequence generation removes boundary discontinuities, it inevitably alters originally unedited regions, which is undesirable for localized editing. Accordingly, we evaluate both our method and the baselines over the full sequence.

\Tref{tab:subquan} summarizes the quantitative results. Although lip-sync accuracy (LSE-D/C) improves slightly for most baselines compared to the main paper, identity preservation (IDSIM) degrades, indicating drift in the subject’s appearance during full-sequence generation. We further compute LPIPS on the unedited portions by comparing the synthesized frames with the corresponding original frames. FacEDiT achieves the lowest LPIPS, showing better preservation of unedited content, whereas all baselines yield substantially higher LPIPS, confirming that portrait animation models fail to preserve the unedited regions and noticeably modify them. Overall, this experiment highlights the necessity of a dedicated talking head video editing model.

\begin{table}[!t]
    \caption{\textbf{Comparison on talking head video editing.} Note that the compared models are designed for speech-driven portrait animation and thus serve only as baselines, as they are not directly applicable to editing. We highlight the best results in \textbf{bold} and \uline{underline} the second best.}
    \label{tab:subquan}
    \centering
    \resizebox{0.78\linewidth}{!}{\begin{NiceTabular}{lccccc}
    \toprule
    Method & LSE-D ($\downarrow$)& LSE-C ($\uparrow$) & IDSIM ($\uparrow$) & FVD ($\downarrow$) & LPIPS ($\downarrow$)\\
    \cmidrule{1-6}
    V-Express~\cite{vexpress}& 7.718& \textbf{7.264}& 0.756&196.974&0.426\\
    AniPortrait~\cite{aniportrait}& 11.421& 2.651& \uline{0.847}&151.998&0.330\\
    EchoMimic~\cite{chen2025echomimic}& 9.511& 4.890& 0.832&143.894&0.350\\
    EchoMimicV2~\cite{echomimicv2}& 10.136& 3.819& 0.817&124.219&0.347\\
    SadTalker~\cite{sadtalker}& \uline{7.578}& 7.119& 0.779&132.646&0.324\\
    Hallo~\cite{hallo}& 8.210& 6.591& 0.842&116.429&0.315\\
    Hallo2~\cite{hallo2}& 8.191& 6.546& 0.843&110.768&0.317\\
    Hallo3~\cite{cui2025hallo3}& 8.805& 6.368& 0.796&\uline{106.796}&0.334\\
    KeyFace~\cite{keyface}& 9.258& 5.289& 0.796&129.082&\uline{0.307}\\
    \hline
    \rowcolor{blue!5}
    Ours& \textbf{7.523}& \uline{7.218} & \textbf{0.971}&\textbf{44.766}&\textbf{0.126}\\
    \bottomrule
    \end{NiceTabular}}
\end{table}

\begin{table}[!t]
\caption{\textbf{Comparison of inference time.} We measure the inference time of each model when generating a talking head video from the same 15 seconds of speech.}
    \label{tab:inference_time}
    \centering
    \begin{minipage}{0.45\linewidth}
        \centering
        \resizebox{\linewidth}{!}{
        \begin{NiceTabular}{lc}
            \toprule
            Method & Inference (seconds) \\
            \cmidrule{1-2}
            \rowcolor{blue!5}
            Ours & 73.787\\
            V-Express~\cite{vexpress} & 639.694\\
            AniPortrait~\cite{aniportrait} & 280.817 \\
            EchoMimic~\cite{chen2025echomimic} & 808.065 \\
            EchoMimicV2~\cite{echomimicv2} & 512.349\\
            \bottomrule
        \end{NiceTabular}}
    \end{minipage}
    \hfill
    \begin{minipage}{0.4\linewidth}
        \centering
        \resizebox{\linewidth}{!}{
        \begin{NiceTabular}{lc}
            \toprule
            Method & Inference (seconds)  \\
            \cmidrule{1-2}
            SadTalker~\cite{sadtalker} & 177.455 \\
            Hallo~\cite{hallo}& 1029.339\\
            Hallo2~\cite{hallo2}& 1056.977\\
            Hallo3~\cite{cui2025hallo3}& 2570.673\\
            KeyFace~\cite{keyface}& 372.119\\
            \bottomrule
        \end{NiceTabular}}
    \end{minipage}
\end{table}

\begin{figure*}[!t]
    \centering
    \includegraphics[width=1\linewidth]{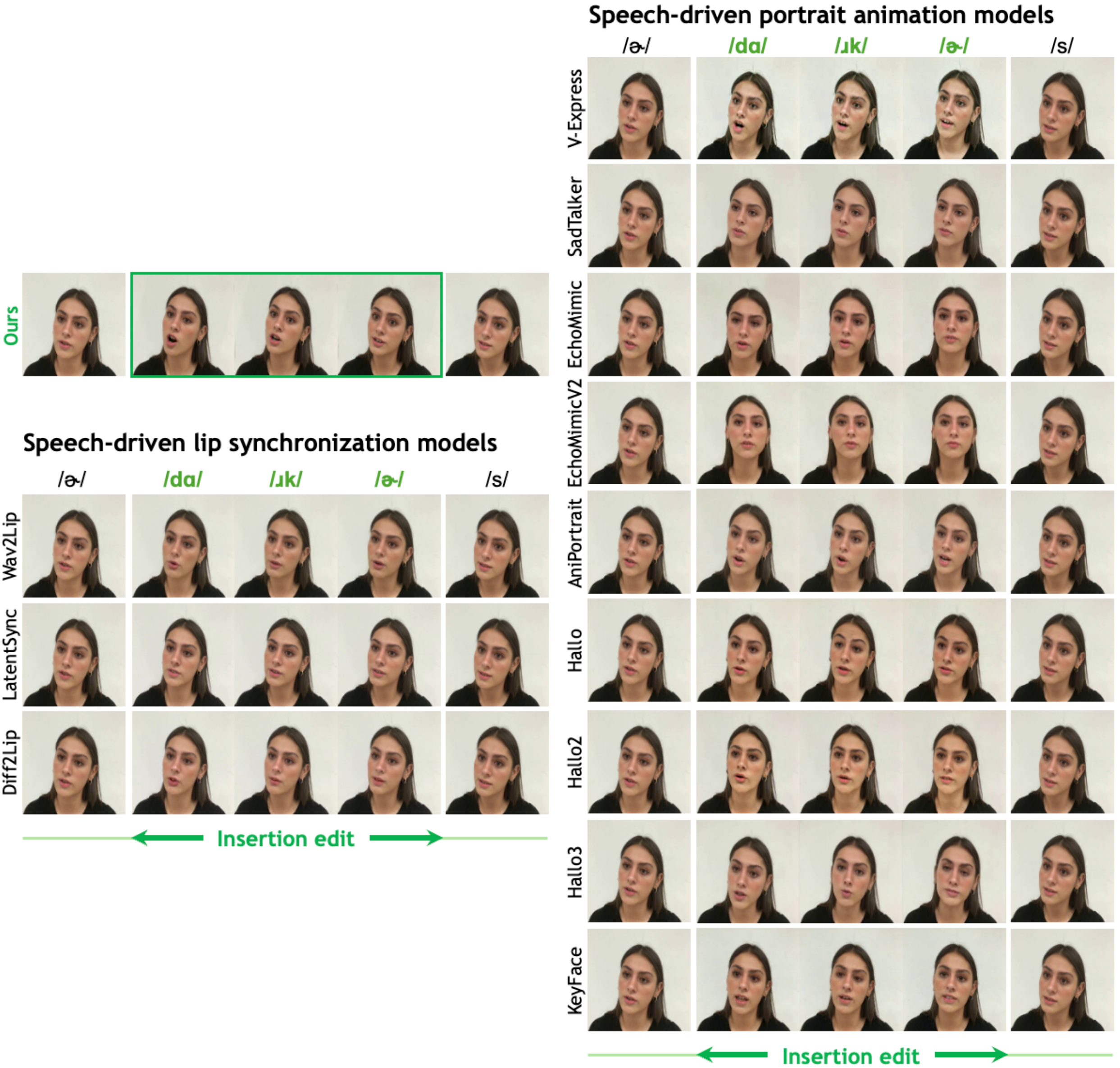}
    \caption{\textbf{Qualitative comparison.} Our method produces accurate lip synchronization with the edited speech while preserving identity and maintaining smooth transitions between edited and unedited frames compared to baseline models.}
    \label{sup:qual}
    
\end{figure*}

\paragraph{Inference time}
Leveraging the efficient facial motion representation of LivePortrait~\cite{guo2024liveportrait}, our model achieves substantially faster inference compared to prior work. \Tref{tab:inference_time} summarizes the inference time for generating a 15-second video on a single RTX A6000 GPU (for Hallo3, which requires more than 48GB of memory, we instead benchmark on an A100 GPU). Our model achieves up to \emph{2.4$\times$ faster} inference than the second-fastest portrait animation baseline, SadTalker~\cite{sadtalker}, and up to \emph{35$\times$ faster} than Hallo3~\cite{cui2025hallo3}. These results highlight not only the superior performance of our method in both objective and subjective evaluations, but also its remarkable efficiency, making it well-suited for both talking head video editing and speech-driven portrait animation.

\paragraph{Qualitative comparison}
We showcase synthesized results from all models, including those partially shown in the main paper, on a talking head video editing sample (\Fref{sup:qual}). Overall, our method produces accurate lip synchronization consistent with the edited transcript. Moreover, it preserves identity and maintains smooth transitions between edited and unedited frames, whereas the baseline models often result in blurred lip motion, boundary discontinuities, or identity drift. These results highlight the effectiveness of our approach for talking head video editing.

\begin{table}[t]
    \centering
    \caption{\textbf{Motion smoothness analysis.}
    RMS velocity and acceleration of predicted facial-motion trajectories
    with and without the temporal smoothness loss.}
    \label{tab:smoothness_analysis}
    \resizebox{0.5\linewidth}{!}{%
    \begin{tabular}{lcc}
        \toprule
        \textbf{Variant}
        & \textbf{Velocity $\downarrow$}
        & \textbf{Acceleration $\downarrow$} \\
        \midrule
        w/o $\mathcal{L}_{\mathrm{TS}}$
        & 0.001983 & 0.001924 \\
        w/ $\mathcal{L}_{\mathrm{TS}}$
        & \textbf{0.001872} & \textbf{0.001671} \\
        \midrule
        Relative reduction
        & 5.6\% & 13.2\% \\
        \bottomrule
    \end{tabular}}
\end{table}

\paragraph{Analysis of temporal smoothness loss}
We further analyze the effect of the temporal smoothness loss
$\mathcal{L}_{\mathrm{TS}}$ on the predicted facial-motion trajectories.
We compare the same ablation configurations with and without
$\mathcal{L}_{\mathrm{TS}}$, and compute RMS velocity and acceleration from the first- and second-order temporal differences of the predicted motion
latents over FacEDiTBench.
As shown in \Tref{tab:smoothness_analysis}, adding
$\mathcal{L}_{\mathrm{TS}}$ reduces the RMS velocity from 0.001983 to
0.001872 ($-5.6\%$) and the RMS acceleration from 0.001924 to
0.001671 ($-13.2\%$).
These reductions indicate smoother motion dynamics and are consistent with
the improved boundary-continuity metrics in the main ablation study.

\section{Additional implementation details}\label{app:impl_details}
\paragraph{Facial motion representation}
For facial motion representation, we adopt the motion latents from LivePortrait~\cite{guo2024liveportrait}, a portrait animation framework that can animate a source image using a driving signal (\eg, video or motion).
For each image frame, the appearance feature $h$ and the motion latent $q$ 
are extracted through independent extractors. The motion latent is defined as $q=s\cdot(q_c R+\delta)+\tau$, where $q_c \in \mathbb{R}^{21\times3}$ denotes canonical keypoints, $R \in \mathbb{R}^{3\times3}$ represents head rotation, $\delta \in \mathbb{R}^{21\times3}$ captures expression deformation, $\tau \in \mathbb{R}^{1\times3}$ is translation, and $s$ is a scale factor.
Given source appearance features $h_s$, source keypoints $q_s$, and driving keypoints $q_d$, the warping module produces a warped feature that is decoded by the LivePortrait decoder to render the target animated frame.

Among the motion components, we use $[\delta, R, \tau] \in \mathbb{R}^{25\times3}$ (concatenated across channels), which encode the essential facial dynamics for animation. We flatten these into a vector $f \in \mathbb{R}^{75}$ as our facial motion representation. 
This choice offers several practical advantages: (i) the representation is compact, structured, and easy to control; and (ii) the disentanglement between appearance and motion helps preserve identity, since the appearance feature remains fixed while only the relevant motion components are modified for talking head video synthesis.

\subsection{Training} 
During training, we apply random masking to the facial motion sequence to perform motion infilling. Specifically, we randomly sample the temporal location and length of the masked span, covering 70\%--100\% of the sequence.
The channel-wise mask follows one of two predefined patterns: a full-motion mask hides all facial motion channels within the selected span, whereas a lip-only mask hides only the lip-related channels while preserving the remaining expression, head rotation, and translation
components.
Thus, random masking refers to the temporal span, rather than to randomly selecting individual motion dimensions.
For the attention bias, we use a window length of $w=10$. The extracted speech features are at 50 Hz, while the video is at 25 FPS; therefore, we interpolate the facial motion features to 50 FPS so that the sampling rates of speech and facial motion are aligned.

\subsection{Inference}
As described in the main paper, our unified facial motion infilling formulation $p(\mathbf{F}_{\mathbf{M}} \mid \mathbf{F}_{\bar{\mathbf{M}}}, \mathbf{A})$, where $\mathbf{F}_{\mathbf{M}}$ denotes the masked motion, $\mathbf{F}_{\bar{\mathbf{M}}}$ denotes the visible motion, $\mathbf{A}$ denotes the speech feature, and $\mathbf{M}$ is the mask, can be instantiated for talking head video editing, speech-driven portrait animation, and speech-driven lip synchronization through task-specific mask definitions.

\paragraph{Talking head video editing}
Let the edited segment correspond to the frame range $[t_1, t_2]$. The temporal mask is defined as:
\[
M_t = \begin{cases} 1 & t_1 \le t \le t_2 \\ 0 & \text{otherwise} \end{cases}.
\]
In this setup, the model infills the masked motion within the edited interval while conditioning on the surrounding frames and speech. Thus, editing is treated as a temporal motion infilling task.

\paragraph{Speech-driven portrait animation}
Let the observed source motion be given up to frame $T$. The mask for speech-driven portrait animation is then:
\[
M_t = \begin{cases} 0 & t \le T \\ 1 & t > T \end{cases}.
\]
The model generates the subsequent motion sequence conditioned on the speech and the initial observed state. Consequently, portrait animation corresponds to sequence completion via motion infilling.

\paragraph{Speech-driven lip synchronization}
Following \cite{kwon2025jam}, the LivePortrait motion latent can be decomposed into fine-grained components:
\[
f_t = (f_t^{rest}, f_t^{lip}, f_t^{pose}, f_t^{trans}),
\]
where $f_t^{rest}$ represents non-lip facial expressions, and $f_t^{lip}, f_t^{pose}, \text{ and } f_t^{trans}$ correspond to lip motion, head pose, and translation, respectively. For speech-driven lip synchronization, we define a channel-wise mask:
\[
M_{t, rest} = 0, \quad M_{t, lip} = 1, \quad M_{t, pose} = 0, \quad M_{t, trans} = 0.
\]
Under this configuration, the model reconstructs lip motion conditioned on speech and the unmasked motion components while preserving the remaining parameters. Thus, lip synchronization is formulated as a channel-wise motion infilling task.

\subsection{Frame resampling module}\label{app:resampling}
A key challenge is that the edited region may differ in length from the original segment, requiring frame resampling before final video rendering. Naive duplication or interpolation for upsampling can introduce artifacts, while downsampling often causes severe aliasing. To address this, we design a face-aware resampling module based on dense optical flow and soft interpolation, which adjusts the RGB frame sequence to the edited duration while preserving coherent motion and visual continuity. The resampled frames, together with the motion predicted by FacEDiT, are then passed to the LivePortrait decoder~\cite{guo2024liveportrait} to render the final edited video.

Given an input sequence of $N$ frames, we generate $M$ output frames by retaining the first and last frames and synthesizing the remaining $(M-2)$ intermediate frames. Let $I_0$ and $I_L$ denote the endpoints, and let $t \in (0,1)$ be the normalized timestamp of an intermediate frame. We synthesize $I_t$ by (1) estimating bidirectional flows between $I_0$ and $I_L$, (2) warping both endpoints toward time $t$, and (3) blending the warped images using a softmax-weighted forward-splatting formulation~\cite{niklaus2020softmax}. This endpoint-based strategy reduces aliasing during downsampling and avoids abrupt frame repetition during upsampling.

\paragraph{Endpoint flows}
First, we estimate the dense optical flow between the endpoints using RAFT~\cite{teed2020raft} as:
\[
F_{0\rightarrow L}=\text{RAFT}(I_0, I_L), \qquad
F_{L\rightarrow 0}=\text{RAFT}(I_L, I_0).
\]
To approximate motion at time $t$, we scale the flows as
\[
F_{0\rightarrow t}=t\,F_{0\rightarrow L}, \qquad
F_{L\rightarrow t}=(1-t)\,F_{L\rightarrow 0},
\]
and warp $I_0$ and $I_L$ accordingly to obtain $I_0^t$ and $I_L^t$.

\paragraph{Soft forward--backward reliability}
To account for occlusions and inaccurate flow, we compute soft forward--backward (FB) consistency weights:
\begin{align*}
w_0=\exp(-\beta\,\text{FB}(F_{0\rightarrow L}, F_{L\rightarrow 0})),\\
w_L=\exp(-\beta\,\text{FB}(F_{L\rightarrow 0}, F_{0\rightarrow L})),
\end{align*}
where $\operatorname{FB}(x)=\lVert F_{a\to b}(x) + F_{b\to a}\bigl(x + F_{a\to b}(x)\bigr)\rVert_2$, with $x$ a pixel coordinate, measures the forward-backward flow inconsistency. A small floor value is applied to avoid zero weights and ensure stable blending.

\paragraph{Face-aware reliability (MediaPipe)}
Because talking head videos exhibit structured facial motion, we incorporate a face-aware spatial prior to emphasize reliable regions during blending. We use MediaPipe Face Mesh~\cite{lugaresi2019mediapipe} to detect dense facial landmarks on $I_0$ and $I_L$, extract the facial contour, and rasterize it into binary face masks $\text{ROI}_0$ and $\text{ROI}_L$. These masks amplify reliability within the face region:
\[
w_0' = w_0 \big(0.8 + 0.2\,\text{ROI}_0\big), \qquad
w_L' = w_L \big(0.8 + 0.2\,\text{ROI}_L\big),
\]
encouraging smooth motion around the face.

\paragraph{Softmax forward splatting}
The intermediate $(M-2)$ frames are synthesized via softmax-weighted forward splatting of the warped endpoints:
\[
\alpha_0 = (w_0')^{\gamma}(1-t), \qquad
\alpha_L = (w_L')^{\gamma} t,
\]
\[
I_t =
\frac{
\alpha_0 \cdot \text{splat}(I_0^t) +
\alpha_L \cdot \text{splat}(I_L^t)
}{
\alpha_0 + \alpha_L + \varepsilon
},
\]
where $\gamma$ is the softmax sharpness exponent for blending and
$\text{splat}(I)(x)= \sum_{y} I(y)\, \omega(y \rightarrow x),$
with $\omega(y \rightarrow x)$ denoting the bilinear weight by which pixel $y$ contributes to location $x$ after being warped by the flow. Forward splatting distributes each source pixel to the target frame using flow-aligned bilinear weights, providing dense coverage even under large motion.

\paragraph{Hole-aware fallback}
If the accumulated splat weight falls below a threshold, we fall back to the nearest endpoint warp.
\[
I_t(x)=
\begin{cases}
I_0^t(x), & t < 0.5,\\[2pt]
I_L^t(x), & t \ge 0.5.
\end{cases}
\]

\paragraph{Result}
The resampled sequence
\[
[I_0,\; I_{t_1},\; I_{t_2},\,\dots,\, I_{t_{M-2}},\; I_L]
\]
contains all interior frames synthesized via face-aware softmax forward splatting. This produces temporally smooth, identity-consistent, and hole-free transitions, which are crucial for the later stages of facial animation rendering.

\begin{figure*}[!t]
    \centering
    \includegraphics[width=0.9\linewidth]{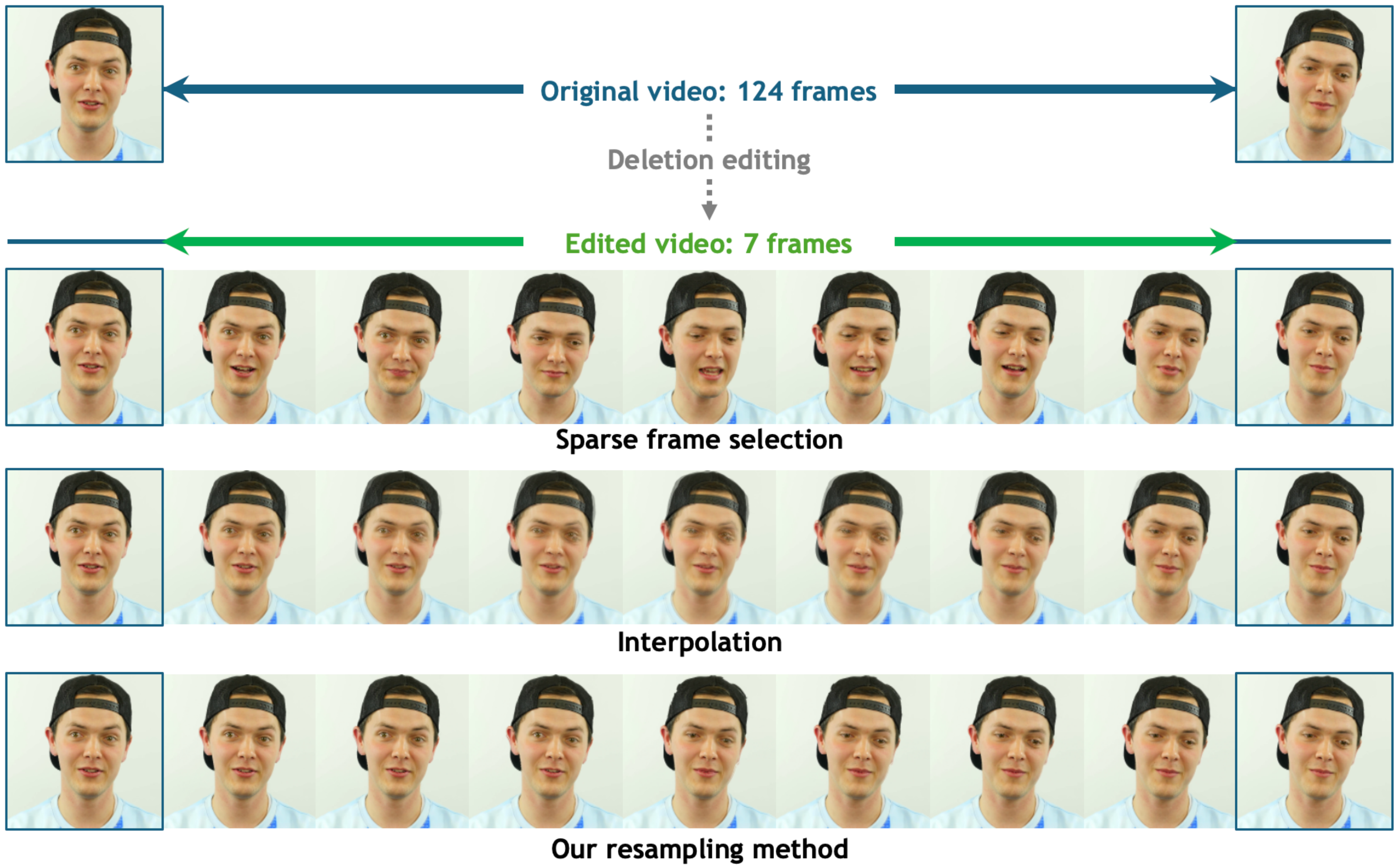}
    \caption{\textbf{Qualitative comparison of frame resampling methods.} When downsampling a video (\eg, from 124 frames to 7), an effective resampling strategy is required. Sparse frame selection introduces aliasing, while interpolation between only the first and last frames produces noticeable blur. In contrast, our resampling method produces smoother transitions with significantly fewer artifacts.}
    \label{fig:resample}
\end{figure*}

\begin{table}[!t]
    \caption{\textbf{Quantitative comparison of resampling methods.} Using our frame resampling approach during downsampling improves lip-sync accuracy and boundary continuity compared to sparse frame selection and interpolation-based methods.}
    \label{tab:resample}
    \centering
    \resizebox{0.8\linewidth}{!}{\begin{NiceTabular}{lccccc}
    \toprule
    Method & LSE-D ($\downarrow$)& LSE-C ($\uparrow$) & IDSIM ($\uparrow$) &  P$_{\text{continuity}}$ ($\downarrow$) & M$_{\text{continuity}}$ ($\downarrow$)\\
    \cmidrule{1-6}
    Sparse selection& 7.769&5.137 &\textbf{0.968} &3.609&1.587\\
    Interpolation&\uline{7.716} & \uline{5.215}& 0.954&\uline{3.139}&\uline{0.829}\\
    \rowcolor{blue!5}
    Our resampling&\textbf{7.336} &\textbf{5.588} &\uline{0.960}&\textbf{3.118}&\textbf{0.825}\\
    \bottomrule
    \end{NiceTabular}}
\end{table}

Figure~\ref{fig:resample} compares different downsampling strategies for reducing 124 frames to 7. A simple baseline is sparse frame selection, which samples frames at evenly spaced intervals; however, this causes strong aliasing and abrupt temporal changes. Another approach is linear interpolation between the endpoint frames, but pixel-space interpolation across such a long temporal gap introduces noticeable blur. In contrast, our resampling module produces smooth motion with far fewer artifacts. These observations are supported by \Tref{tab:resample}, where we report quantitative results on editing samples that require downsampling. We find that our resampling method improves lip-sync accuracy as well as boundary continuity compared to both sparse sampling and interpolation-based approaches.

\section{Details on the experimental setup}\label{app:exp_setup}
\subsection{Comparison with speech-driven portrait animation models}\label{sec:comp_portrait}
We compare against nine speech-driven portrait animation models on both the talking head video editing and portrait animation tasks. For the editing setting, we provide only the edited speech segment (excluding unedited portions) to each model, generate the corresponding video segment, and stitch it back into the original video. Note that these models are designed for portrait animation rather than editing, and thus serve only as baselines for the editing task.

\paragraph{V-Express~\cite{vexpress}}
V-Express adopts a progressive training strategy that balances strong and weak control signals. A latent diffusion backbone is extended with conditioning modules for reference images and audio cues, while V-Kps pose inputs are encoded by a lightweight guider and injected directly into the latent space. Conditional dropout prevents over-reliance on strong conditions, enabling effective audio-driven motion. This design produces identity-consistent, pose-aligned, and audio-synchronized portrait videos.

\paragraph{AniPortrait~\cite{aniportrait}}
AniPortrait maps audio to 3D facial structures and projects them into 2D landmarks, which serve as a robust intermediate representation for a diffusion-based generator. By conditioning on these audio-driven landmarks and a reference portrait image, the model synthesizes temporally coherent frames with strong identity preservation. The explicit 3D geometry offers controllable motion and enables natural expressions as well as applications, such as facial motion editing.

\paragraph{EchoMimic~\cite{chen2025echomimic}}
EchoMimic jointly leverages audio cues and facial landmarks within a unified diffusion framework. Audio and landmark features are encoded separately and injected into the denoising process, allowing the model to operate with audio only, landmarks only, or both. This dual-conditioning setup stabilizes motion prediction under weak audio signals while maintaining geometric consistency informed by landmarks. The resulting animations preserve identity and offer fine-grained control of facial motion.

\paragraph{EchoMimicV2~\cite{echomimicv2}}
EchoMimicV2 extends generation to the half-body setting through a streamlined conditioning strategy. An audio–pose dynamic harmonization module reduces dependence on detailed pose inputs while broadening audio control from local lip motion to full-body dynamics. Minimal structural cues, such as a single-frame hand-pose estimate, guide expressive upper-body movements. Phase-specific denoising objectives further improve stability. For fair comparison, we provide hand pose extracted from the first frame using the authors' implementation.

\paragraph{SadTalker~\cite{sadtalker}}
SadTalker estimates 3DMM motion coefficients from audio, including head pose and expression parameters. The predicted motion is transformed into an unsupervised 3D keypoint space and rendered through a 3D-aware generator for strong identity preservation. By modeling the relationship between audio and various motion components, the method produces stable lip movements and expressive head dynamics with controllable animation. 

\paragraph{Hallo~\cite{hallo}}
Hallo uses a hierarchical audio–visual cross-attention mechanism that aligns audio features with specific facial regions. Region-specific masks for lips, expressions, and head pose are applied within the same latent resolution, enabling fine-grained motion control while maintaining reference-image appearance. This hierarchical structure enhances temporal coherence and motion diversity while preserving identity.

\paragraph{Hallo2~\cite{hallo2}}
Hallo2 employs incremental motion synthesis with patch-drop augmentation to improve long-range consistency. Motion tokens are predicted autoregressively and aligned with audio cues through a transformer backbone. Patch-drop increases robustness to missing temporal information, while hierarchical cross-attention fuses audio, motion, and spatial features into a unified representation.

\paragraph{Hallo3~\cite{cui2025hallo3}}
Hallo3 uses a video diffusion transformer that models global spatiotemporal relationships across frames. Appearance cues from the reference image, audio-driven motion signals, and text-guided modulation are integrated to produce expressive, identity-preserving facial dynamics. Because Hallo3 requires a text prompt that indicates the appearance at inference time, we use a fixed neutral prompt (``A person talking'') for all comparisons.

\paragraph{KeyFace~\cite{keyface}}
KeyFace enhances temporal coherence via a keyframe-driven synthesis approach. In stage 1, a sparse set of key motion frames is predicted, followed by stage 2 interpolation of intermediate frames using a U-Net denoiser conditioned on audio features. This keyframe--interpolation design stabilizes identity, reduces temporal drift, and produces smooth, lip-synced motion over long sequences. Progressive conditional dropout further enhances robustness during training. In the main paper, we also report a KeyFace variant: boundary frames from the original video (corresponding to the edit boundaries) are used as stage-1 keyframes, enabling stage-2 interpolation to better preserve boundary continuity than the original setup.

\subsection{Comparison with speech-driven lip synchronization models}\label{sec:comp_lip_sync}
We compare three speech-driven lip synchronization models on both talking head video editing and lip synchronization, and additionally include two variants adapted from portrait animation models as lip synchronization baselines. For editing, we resample the original video to match the edited speech length, use it as the driving video, and modify the lower-face region according to the edited speech. As these models are designed for lip synchronization, they serve only as editing baselines.

\paragraph{LatentSync~\cite{li2024latentsync}}
LatentSync is an end-to-end lip-sync framework built on an audio-conditioned latent diffusion model that operates in the latent space of Stable Diffusion, avoiding explicit intermediate motion representations such as landmarks or 3D models. To improve temporal consistency across generated frames, it introduces TREPA (Temporal REPresentation Alignment), which aligns temporal representations of generated and ground-truth frame sequences extracted by a large-scale self-supervised video model, serving as an additional temporal regularizer without adding model parameters.

\paragraph{Diff2Lip~\cite{mukhopadhyay2024diff2lip}}
Diff2Lip formulates lip synchronization as an audio-conditioned diffusion inpainting problem over the mouth region. It takes three inputs: (i) a masked input frame to provide pose/context, (ii) a reference frame to provide identity and mouth texture, and (iii) an audio frame/window to drive lip shape; the diffusion model then generates the lip-synced mouth region conditioned on these cues.

\paragraph{Wav2Lip~\cite{prajwal2020lip}}
Wav2Lip uses a CNN encoder--decoder generator with three components: an identity encoder (encoding a reference frame together with a pose prior formed by masking the lower half of the target face), a speech encoder for the audio segment, and a face decoder to synthesize the output frame. A key module is a pretrained lip-sync expert discriminator (SyncNet-style), which provides a synchronization loss that penalizes audio--visual mismatch; the generator can also be trained with a separate visual-quality discriminator to improve realism.

\paragraph{Variants of SadTalker~\cite{sadtalker} and AniPortrait~\cite{aniportrait}}
These variants adapt SadTalker and AniPortrait, originally designed for speech-driven portrait animation, to the lip synchronization setting. The setup matches other lip synchronization models: a driving video provides head pose and other non-lip motion. Since both SadTalker and AniPortrait leverage 3D structure as intermediate features, we replace the non-lip components with those from the driving video and use the predicted lip-related 3D features, resulting in lip synchronization variants.


\subsection{Details on the objective metrics}
\paragraph{LSE-D and LSE-C}
To evaluate lip synchronization accuracy, we measure the audio–visual alignment between lip movements and speech using SyncNet~\cite{syncnet}, which learns joint representations of audio and visual signals. SyncNet yields two metrics: Lip Sync Error–Distance (LSE-D) and Lip Sync Error–Confidence (LSE-C). LSE-D computes the Euclidean distance between audio and visual embeddings, with lower values indicating better synchronization. LSE-C is a confidence score derived from these distances, where higher values denote more reliable audio–visual alignment.

\paragraph{Fréchet Video Distance (FVD)}
To assess perceptual realism in synthesized talking head videos, we employ FVD~\cite{fvd}, which measures the distributional similarity between generated and real videos. Following standard practice, we extract video features using a pre-trained I3D network and compute the Fréchet distance between the feature distributions of synthesized and ground-truth videos, capturing both temporal dynamics and overall visual quality.

\paragraph{LPIPS}
LPIPS~\cite{zhang2018unreasonable} quantifies frame-level perceptual similarity using image features extracted from a pre-trained network. In talking head video synthesis, LPIPS reflects how well identity, facial structure, and appearance are preserved. We compute LPIPS frame-by-frame by comparing matching timestamps from the synthesized and reference videos.

\begin{figure*}[!t]
    \centering
    \includegraphics[width=1\linewidth]{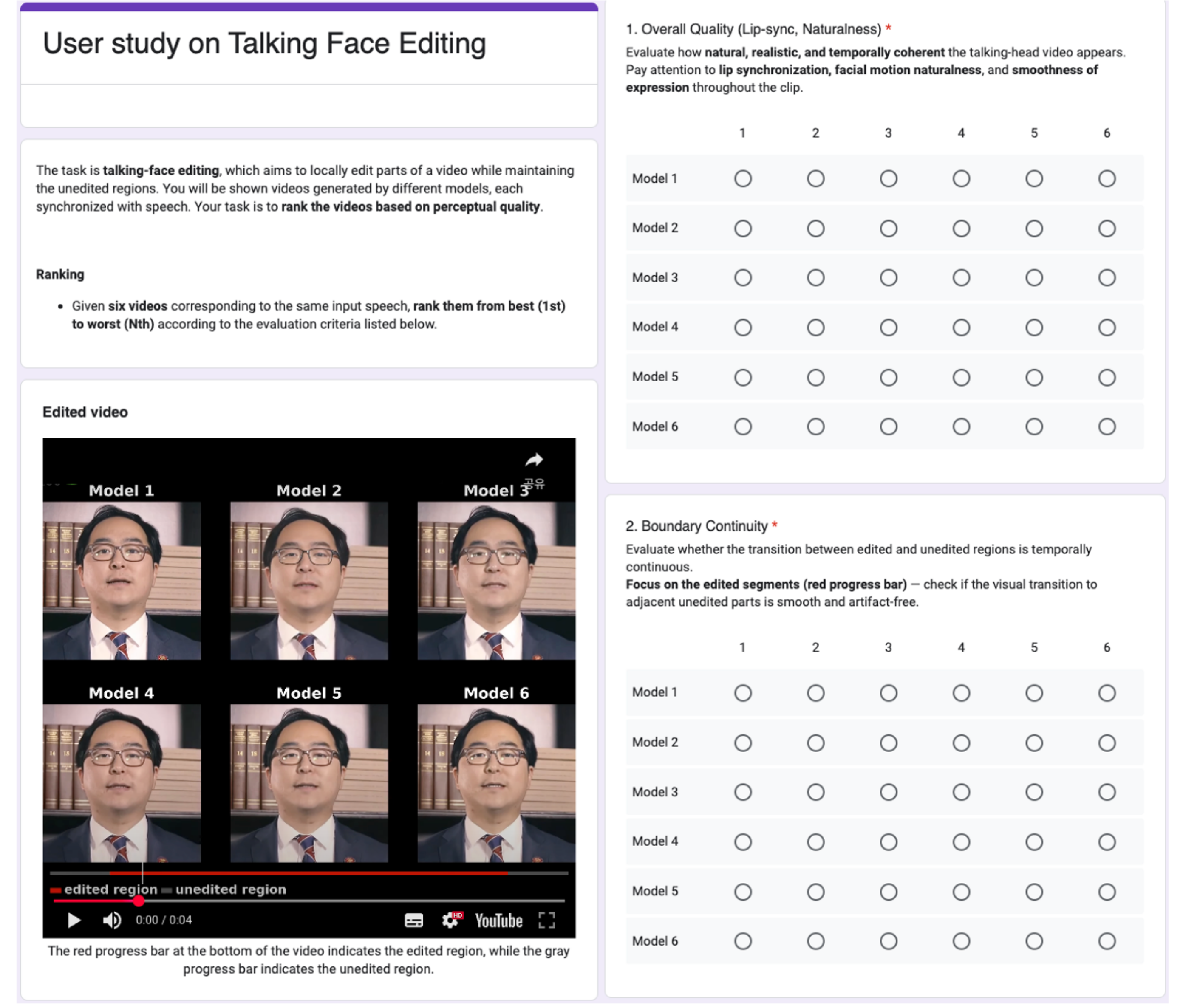}
    \vspace{-3mm}
    \caption{\textbf{Instruction and example for the human study.} We provide a sample template used in the talking head video editing evaluation.}
    \label{fig:user}
    
\end{figure*}

\subsection{Human evaluation}\label{app:human}
We conduct a human study to compare FacEDiT with baselines from speech-driven portrait animation and lip synchronization. We select five recent, high-performing portrait animation models~\cite{sadtalker,hallo,hallo2,cui2025hallo3,keyface} from nine candidates and three lip synchronization models~\cite{li2024latentsync,mukhopadhyay2024diff2lip,prajwal2020lip}. We form two independent comparison pools, one for each model category, and evaluate 20 synthesized videos per method with 30 participants.
For talking head video editing, participants separately rank the methods in each pool by overall quality and boundary continuity. Overall quality considers lip synchronization, facial motion naturalness, and expression smoothness, while boundary continuity measures the temporal smoothness between edited and unedited regions. A red progress bar marks the edit region to help identify the modified segment; the instructions and an example are shown in \Fref{fig:user}. We additionally evaluate the corresponding generation task in each pool, \ie, portrait animation or lip synchronization, using overall quality only. Rankings are computed independently within each comparison pool.

\section{Theoretical interpretation of facial motion infilling}
\label{app:theory}

We provide a theoretical interpretation of FacEDiT as a mask-conditional generative model. We first show that multiple talking head synthesis tasks can be expressed as facial motion infilling, and then analyze how random-mask training supports other tasks within a unified model.

Let $\mathbf{F}=(\mathbf{f}_1,\ldots,\mathbf{f}_T)\in\mathbb{R}^{T\times C}$
denote a sequence of facial motion latents, $\mathbf{A}$ denote speech
features, and $\mathbf{M}\in\{0,1\}^{T\times C}$ denote a binary spatiotemporal mask,
where $\mathbf{M}_{t,j}=1$ indicates that the corresponding motion entry is masked.
The masked and visible motion entries are denoted by
$\mathbf{F}_{\mathbf{M}}$ and $\mathbf{F}_{\bar{\mathbf{M}}}$, respectively, \ie,
$\mathbf{F}_{\mathbf{M}}=\mathbf{M}\odot\mathbf{F}$ and
$\mathbf{F}_{\bar{\mathbf{M}}}=(1-\mathbf{M})\odot\mathbf{F}$, where $\odot$ denotes element-wise multiplication.
Our facial motion infilling aims to model the conditional probability distribution
\begin{equation}
    p(\mathbf{F}_{\mathbf{M}}\mid \mathbf{F}_{\bar{\mathbf{M}}},\mathbf{A}),
    \label{eq:app_infilling}
\end{equation}
\ie, the conditional distribution of missing facial motion given visible motion
context and speech.

\subsection{Task unification by mask specification}
\label{app:task_unification}

We first show that the probability distribution in Eq.~\eqref{eq:app_infilling} is general, such that it can represent talking head video editing, portrait animation, and lip synchronization as special cases according to the mask.

\begin{proposition}[Task unification via motion infilling]
\label{thm:task_unification}
Talking head video editing, speech-driven portrait animation,
and speech-driven lip synchronization can all be expressed as facial motion
infilling problems by choosing different masks \(\mathbf{M}\).
\end{proposition}

\begin{proof}
The proof proceeds by showing that each task can be reformulated as a special case of conditional motion infilling under an appropriate choice of mask $\mathbf{M}$.
Specifically, we interpret the mask as specifying which motion components are observed and which components must be generated.
Different tasks correspond to different masking strategies, including temporal masking and channel-wise masking.

For talking head video editing, let the edited interval be $[t_1,t_2]$. We define
\[
    \mathbf{M}_t=
    \begin{cases}
        1, & t_1\leq t\leq t_2,\\
        0, & \text{otherwise}.
    \end{cases}
\]
The model predicts motion only inside the edited interval while conditioning on
surrounding frames and edited speech. Thus, editing is temporal motion infilling.

For portrait animation, suppose an initial motion prefix up to frame $T_0$ is
observed. We define
\[
    \mathbf{M}_t=
    \begin{cases}
        0, & t\leq T_0,\\
        1, & t>T_0.
    \end{cases}
\]
The model then predicts the remaining motion sequence conditioned on the speech
and the observed prefix. Thus, portrait animation is sequence completion via
motion infilling.

For lip synchronization, decompose the motion latent as
\[
    \mathbf{f}_t=
    \left(
    \mathbf{f}^{\mathrm{rest}}_t,
    \mathbf{f}^{\mathrm{lip}}_t,
    \mathbf{f}^{\mathrm{pose}}_t,
    \mathbf{f}^{\mathrm{trans}}_t
    \right).
\]
We use the channel-wise mask
\[
    \mathbf{M}_{t,\mathrm{rest}}=0,\qquad
    \mathbf{M}_{t,\mathrm{lip}}=1,\qquad
    \mathbf{M}_{t,\mathrm{pose}}=0,\qquad
    \mathbf{M}_{t,\mathrm{trans}}=0.
\]
The model reconstructs only lip motion while preserving pose, translation, and
non-lip facial motion. Thus, lip synchronization is channel-wise motion infilling.

Therefore, all three tasks are instances of the same conditional generation problem
in Eq.~\eqref{eq:app_infilling}, with different choices of $\mathbf{M}$.
\end{proof}

This result suggests that a single conditional generative model can perform editing,
animation, and lip synchronization within the same probabilistic framework by changing
only the masking operator at inference time, \ie, motion infilling as a unified formulation
for multiple talking head generation tasks.
In practice, this does not require architectural modification, which is a favorable property
of our facial motion infilling formulation.

The above proposition shows that the choice of mask $\mathbf{M}$ defines the task structure
through conditional motion infilling, but it does not inform how to obtain the conditional
probability distribution. We next demonstrate how our CFM training induces learning of the
corresponding conditional probability distribution associated with each mask pattern.

\subsection{Mask-conditional flow matching}

We first recall the training setup of our method.
Let $\mathbf{F}_1\sim p_{\mathrm{data}}$ be a ground-truth facial motion sequence and
$\mathbf{F}_0\sim p_0$ be a noise sample. Conditional flow matching uses the linear
interpolation
\begin{equation}
    \mathbf{F}_t=(1-t)\mathbf{F}_0+t\mathbf{F}_1,
    \qquad t\sim\mathcal{U}[0,1],
    \label{eq:app_linear_path}
\end{equation}
whose conditional velocity is
\[
    \mathbf{u}=\mathbf{F}_1-\mathbf{F}_0.
\]
Given a mask $\mathbf{M}$, FacEDiT conditions the velocity model on speech and
visible motion:
\[
    \mathbf{C}_{\mathbf{M}}
    =
    \left(
    \mathbf{A},
    (1-\mathbf{M})\odot\mathbf{F}_1,
    \mathbf{M}
    \right).
\]
We train the model with the following mask-conditional CFM objective function:
\begin{equation}
    \mathcal{L}_{\mathrm{CFM}}(\theta)
    =
    \mathbb{E}_{\mathbf{M}\sim q(\mathbf{M})}
    \mathbb{E}_{\mathbf{F}_1,\mathbf{A},t,\mathbf{F}_0}
    \left[
    \left\lVert
    \mathbf{M}\odot
    \left(
    v_\theta(\mathbf{F}_t,t,\mathbf{C}_{\mathbf{M}})
    -
    (\mathbf{F}_1-\mathbf{F}_0)
    \right)
    \right\rVert_2^2
    \right],
    \label{eq:app_cfm}
\end{equation}
where $q(\mathbf{M})$ is the training mask distribution.
The model trained with Eq.~\eqref{eq:app_cfm} has the following favorable property.

\begin{proposition}[Per-mask optimality under coverage]
\label{thm:permaskopt}
Consider random-mask training with
$\mathbf{M}\sim q(\mathbf{M})$ using the CFM objective in
Eq.~\eqref{eq:app_cfm}.
Let $\mathcal{M}_{\mathrm{task}}$ be the set of masks required for tasks during inference.
Under the assumptions that
\begin{enumerate}
    \item the training mask distribution covers all task masks,
    $q(\mathbf{M}^\star)>0, \forall \mathbf{M}^\star\in\mathcal{M}_{\mathrm{task}}$,
    \item the model class has sufficient capacity to represent the
Bayes-optimal conditional velocity fields for every mask condition in
$\mathcal{M}_{\mathrm{task}}$,
    \item optimization reaches the population-risk minimizer,
\end{enumerate}
the optimal model recovers the Bayes-optimal conditional velocity field over the masked entries for each task mask $\mathbf{M}^\star\in\mathcal{M}_{\mathrm{task}}$.
\end{proposition}

Its proof is provided below.
This proposition suggests that minimizing the CFM objective function learns a mask-conditional velocity field.
This velocity field defines a conditional probability flow transporting the prior distribution
toward the data distribution under the conditioning $\mathbf{C}_{\mathbf{M}}$.
While this objective function does not directly regress the score
$\nabla_{\mathbf{F}_{\mathbf{M}}}\log p(\mathbf{F}_{\mathbf{M}}\mid\mathbf{F}_{\bar{\mathbf{M}}},\mathbf{A})$,
velocity fields and score fields are closely related in probability-flow formulations~\cite{lipman2023flow}.
Therefore, the learned velocity field can be interpreted as implicitly modeling the conditional distribution
$p(\mathbf{F}_{\mathbf{M}}\mid\mathbf{F}_{\bar{\mathbf{M}}},\mathbf{A})$.

According to the proposition, the optimal model is Bayes-optimal for each task mask. This suggests that a single model trained with random masking can cover generation tasks discussed in Proposition~\ref{thm:task_unification}, including talking head video editing, portrait animation, and lip synchronization, among others.

\subsection{Proof of Proposition~\ref{thm:permaskopt}}
Before starting the proof, we introduce several definitions and lemmas that are needed to prove the proposition. 
We first introduce the concept of the Bayes-optimal predictor as follows.

\begin{definition}[Bayes-optimal predictor~\cite{bishop2006pattern}]
Let $x \in \mathcal{X}$ denote an input variable,
$y \in \mathcal{Y}$ denote a target variable,
and $\ell(\hat y,y)$ be a loss function.
The Bayes-optimal predictor is defined as the function
\begin{equation}
f^*(x) = \mathop{\arg\min}\nolimits_{\hat y} \mathbb{E}\bigl[\ell(\hat y,y)\mid x\bigr],\nonumber
\end{equation}
which minimizes the conditional expected risk under the true data distribution.

In particular, under squared loss $\ell(\hat y,y)= \lVert \hat y-y\rVert_F^2$, the Bayes-optimal predictor is given by the conditional expectation
\begin{equation}
f^*(x)=\mathbb{E}[y\mid x].\nonumber
\end{equation}
\end{definition}

We train a generative model with the CFM objective function to approximate the conditional distribution under a mask condition.
To connect Bayes optimality with the CFM objective function, we introduce the following lemma.

\begin{lemma}[Bayes-optimal conditional velocity]
\label{prop:bayes_velocity}
For a fixed mask $\mathbf{M}$, the population minimizer of the masked
CFM loss satisfies
\begin{equation}
\mathbf{M}\odot
v^*(\mathbf{F}_t,t,\mathbf{C}_{\mathbf{M}})
=
\mathbb{E}
\left[
\mathbf{M}\odot(\mathbf{F}_1-\mathbf{F}_0)
\mid
\mathbf{F}_t,t,\mathbf{C}_{\mathbf{M}}
\right].
\label{eq:app_bayes_velocity}
\end{equation}
\end{lemma}

\begin{proof}
For fixed $(\mathbf{F}_t,t,\mathbf{C}_{\mathbf{M}})$, the masked CFM
objective is a squared-loss regression problem with prediction
$\mathbf{M}\odot v_\theta$ and target
$\mathbf{M}\odot(\mathbf{F}_1-\mathbf{F}_0)$.
The squared-loss minimizer is the conditional expectation of this masked
target, which gives Eq.~\eqref{eq:app_bayes_velocity}.
\end{proof}

The above lemma provides a probabilistic interpretation of conditional flow
matching under mask-conditioned generation.
For each mask condition $\mathbf{M}$, the CFM objective learns the
Bayes-optimal conditional velocity field in the least-squares sense.

The learned velocity field defines a conditional probability flow that
transports samples from the prior distribution toward the target motion
distribution conditioned on $\mathbf{C}_{\mathbf{M}}$.
Consequently, minimizing the CFM objective learns the conditional velocity
field associated with the chosen interpolation path under the specified mask.

Although the CFM objective does not explicitly regress the score function
$\nabla_{\mathbf F_{\mathbf M}} \log p(\mathbf F_{\mathbf M} \mid \mathbf F_{\bar{\mathbf M}},\mathbf A),$
probability-flow formulations establish a close relationship between velocity
fields and score-based dynamics~\cite{lipman2023flow}.
Therefore, the learned conditional velocity field can be interpreted as a
Bayes-optimal estimator of the conditional generation dynamics under the
specified masking condition.

While Lemma~\ref{prop:bayes_velocity} characterizes the Bayes-optimal velocity
field for a fixed mask condition, the actual training objective optimizes an
expectation over masking conditions.
Proposition~\ref{thm:permaskopt} shows that, under a coverage condition on the mask
distribution, the resulting predictor remains Bayes-optimal for every
condition considered during inference.

Let $\mathcal{M}_{\mathrm{task}}$ be the set of masks required by the downstream
tasks. Assume that:
\begin{enumerate}
    \item the training mask distribution covers all task masks,
    \[
        q(\mathbf{M}^\star)>0,\qquad
        \forall \mathbf{M}^\star\in\mathcal{M}_{\mathrm{task}};
    \]
    \item the model class is expressive enough to represent the Bayes-optimal
    conditional velocity fields for all masks in $\mathcal{M}_{\mathrm{task}}$
    simultaneously;
    \item optimization reaches the population-risk minimizer.
\end{enumerate}
Then, the optimal model recovers the Bayes-optimal conditional velocity field over the masked entries for each
task mask $\mathbf{M}^\star\in\mathcal{M}_{\mathrm{task}}$:
\begin{equation}
\begin{aligned}
    \mathbf{M}^\star \odot
v_{\theta^*}(\mathbf{F}_t,t,\mathbf{C}_{\mathbf{M}^\star})
=
\mathbb{E}
\left[
\mathbf{M}^\star \odot (\mathbf{F}_1-\mathbf{F}_0)
\mid
\mathbf{F}_t,t,\mathbf{C}_{\mathbf{M}^\star}
\right].
\end{aligned}
\label{eq:app_task_velocity}
\end{equation}

To prove the above proposition, we use the following lemma for generic per-condition Bayes optimality.

\begin{lemma}[Per-condition Bayes optimality under coverage]
\label{lemma:per_cond}
Let $(X,Y,M)\sim p(X,Y,M)$, where $M\in\mathcal{M}$ denotes a condition variable.
Define the condition-specific risk
\[
R_M(\theta) = \mathbb{E}_{(X,Y)\sim p(\cdot,\cdot\mid M)} \big[ \ell(f_\theta(X,M),Y) \big].
\]
Consider a predictor $f_\theta(X,M)$ trained to minimize the expected risk
\begin{equation}
\label{eq:expected_risk}
\min\nolimits_\theta \mathbb{E}_{M \sim p(M)} [R_M(\theta)].
\end{equation}
Under the assumptions that
\begin{enumerate}
    \item the loss $\ell$ is proper, \eg, squared loss,
    \item the predictor class has sufficient capacity to represent the
    Bayes-optimal predictor for every condition in $\mathcal{M}$,
    \item the condition distribution satisfies a coverage condition
    $p(M)>0, \forall M\in\mathcal{M}$,
    where $\mathcal{M}$ includes the set of conditions considered during inference,
\end{enumerate}
any global minimizer $\theta^*$ yields a Bayes-optimal
predictor for every condition $M\in\mathcal{M}$:
\[
f_{\theta^*}(x,m) = \arg\min_{\hat y} \mathbb{E} \big[ \ell(\hat y,Y) \mid X=x,M=m \big].
\]
Moreover, under the squared loss, the Bayes-optimal predictor is
\[
f_{\theta^*}(x,m) = \mathbb{E}[Y\mid X=x,M=m].
\]
\end{lemma}

\begin{proof}
The expected risk in Eq.~\eqref{eq:expected_risk} can be written as
\begin{equation}
\mathbb{E}_{M\sim p(M)}[R_M(\theta)] = \sum\nolimits_{M} p(M) R_M(\theta).
\end{equation}
By the sufficient-capacity assumption, there exists a predictor that
simultaneously attains the Bayes-optimal risk for every condition $M$.
Since $p(M)>0$ for every condition, any global minimizer of the expected risk must therefore attain the minimum conditional risk $R_M$ for every
$M$ in the support of $p(M)$.
Under a proper loss, minimizing $R_M$ yields the Bayes-optimal predictor for that condition.
Furthermore, when the loss is the squared loss, the closed-form Bayes-optimal predictor is given by the conditional mean according to standard regression theory~\cite{bishop2006pattern}.
\end{proof}

\begin{proof}
Proposition~\ref{thm:permaskopt} is a special case of
Lemma~\ref{lemma:per_cond}.

For a fixed mask $\mathbf{M}$, we define the $\mathbf{M}$-conditioned risk
\begin{equation}
    R_{\mathbf{M}}(\theta)
    =
    \mathbb{E}_{\mathbf{F}_1,\mathbf{A},t,\mathbf{F}_0}
    \left[
    \left\lVert
    \mathbf{M}\odot
    \left(
    v_\theta(\mathbf{F}_t,t,\mathbf{C}_{\mathbf{M}})
    -
    (\mathbf{F}_1-\mathbf{F}_0)
    \right)
    \right\rVert_2^2
    \right].
\label{eq:app_mask_risk}
\end{equation}
Then, the CFM objective function in Eq.~\eqref{eq:app_cfm} can be written as
\begin{equation}
    \mathcal{L}_{\mathrm{CFM}}(\theta)
    =
    \mathbb{E}_{\mathbf{M}\sim q(\mathbf{M})}
    \left[
    R_{\mathbf{M}}(\theta)
    \right].
    \label{eq:app_mask_average}
\end{equation}
It is clear that Eq.~\eqref{eq:app_mask_risk} is a special case of Lemma~\ref{lemma:per_cond}.
Furthermore, since $R_{\mathbf{M}}(\cdot)$ is the squared loss, the rest of the proof follows Lemma~\ref{lemma:per_cond}.

For each fixed mask, Lemma~\ref{prop:bayes_velocity} shows that the
Bayes-optimal solution is the corresponding conditional velocity field.
Under the realizability assumption, a single model can represent these solutions for all task masks simultaneously.
Therefore, the population-risk minimizer recovers the Bayes-optimal conditional velocity field for each covered task mask.
In other words, if the training mask distribution sufficiently covers the masking conditions used at inference time, random-mask training allows FacEDiT to support diverse covered task-specific masking patterns within a unified conditional generation framework.
This concludes the proof of Proposition~\ref{thm:permaskopt}.
\end{proof}

\paragraph{Remark}
Due to the realizability assumption, Proposition~\ref{thm:permaskopt}
should be interpreted as an idealized population-risk analysis
of random-mask training.
In practice, for a finite-capacity model, a single shared parameterization simultaneously optimizes multiple mask-conditioned risks.
As a result, minimizing the expected risk $\mathbb{E}_{\mathbf{M}}[R_{\mathbf{M}}(\theta)]$ may involve trade-offs between different masking conditions, and the learned model may not achieve the Bayes optimum for every mask individually.

Despite this, random-mask training is effective.
Under sufficient mask coverage, the training objective encourages the model to learn a shared conditional representation that generalizes across diverse masking conditions.
In this sense, FacEDiT can be viewed as amortizing a family of mask-conditioned generation problems associated with $\mathbf{M}$.

\section{Applications and ethical considerations}\label{app:ethics}
FacEDiT offers both valuable applications and important ethical considerations.
On the positive side, it offers benefits across domains. For individual creators, online content producers, and professional video production teams, FacEDiT enables precise and efficient editing of localized facial-motion segments without requiring costly or time-consuming reshoots. This capability can significantly improve workflows and reduce production overhead in many real-world applications.

However, as with any generative model, FacEDiT also carries risks of misuse. It could be exploited to generate deepfakes or spread misinformation, raising serious ethical and safety concerns. Responsible deployment is therefore essential, supported by safeguards designed to mitigate these risks. In addition, both video and speech data may encode ethnicity-related biases, which can propagate into the model and result in biased or unfair outputs.

To address these concerns, future work should incorporate techniques, such as robust watermarking and deepfake-detection algorithms specifically designed for facial video content. We also plan to pursue this direction to reduce the risk of FacEDiT being misused. Developing reliable detection and watermarking methods can benefit from access to open-source generative models so that researchers can generate training data and properly evaluate their detectors. Thus, open-sourcing FacEDiT is important for helping the community understand potential misuse scenarios and accelerate research on effective mitigation strategies. Incorporating diverse datasets, including a wide range of ethnicities, facial characteristics, and spoken languages, will further help alleviate demographic biases. Establishing clear usage guidelines and ensuring transparent model governance will support safer and more responsible deployment of the technology.

\section{Licenses and terms of existing assets}\label{app:licenses}
We use existing datasets, pretrained models, codebases, and evaluation tools only for research purposes and cite their original creators. We do not redistribute third-party raw datasets or pretrained checkpoints unless their licenses permit redistribution; any released FacEDiTBench materials will be limited to assets and annotations that comply with the original licenses and usage terms.

\begin{itemize}
    \item \textbf{HDTF}: Released under the CC BY 4.0 license.

    \item \textbf{LivePortrait}: Released under the MIT license. Any optional InsightFace pretrained models used with related pipelines are restricted to non-commercial research use.
    
    \item \textbf{VoiceCraft}: Code is released under CC BY-NC-SA 4.0, and model weights are released under the Coqui Public Model License 1.0.0.
    
    \item \textbf{WavLM}: Distributed through Microsoft's UniLM repository, whose root repository is MIT-licensed.
    
    \item \textbf{RAFT}: Released under the BSD-3-Clause license.
    
    \item \textbf{MediaPipe}: Released under the Apache-2.0 license.
    
    \item \textbf{SyncNet}: Released under the MIT license.
    
    \item \textbf{SadTalker}: Released under the Apache-2.0 license.
    
    \item \textbf{EchoMimic / EchoMimicV2}: Released under the Apache-2.0 license.
    
    \item \textbf{Hallo2}: Released under the MIT license, with an S-Lab License 1.0 notice for its high-resolution CodeFormer-derived component.
    
    \item \textbf{KeyFace}: Released under the Apache-2.0 license.
    
    \item \textbf{LatentSync}: Code is released under the Apache-2.0 license, with model weights released under OpenRAIL++.
    
    \item \textbf{Diff2Lip}: Released under the CC BY-NC 4.0 license.

\end{itemize}

\section{LLM usage}\label{app:llm_usage}
We used GPT-4o to generate edited transcript candidates for FacEDiTBench,
with all prompts and outputs manually inspected. Beyond this use, GPT-4o
was used only for grammar and style corrections and was not involved in
method design, experimental analysis, claims, or conclusions.

\end{document}